\pdfoutput=1
\documentclass[11pt]{article}

\usepackage{EMNLP2023}
\usepackage{times}
\usepackage{latexsym}
\usepackage[T1]{fontenc}
\usepackage[utf8]{inputenc}
\usepackage{microtype}
\usepackage{inconsolata}

\usepackage{multirow}
\usepackage{booktabs}
\usepackage{graphicx}
\usepackage{subfigure}
\usepackage{bm}
\usepackage{amsmath}
\usepackage{braket}
\usepackage{color}
\usepackage{colortbl}
\usepackage{tcolorbox}
\usepackage{xcolor}
\usepackage{CJKutf8}
\usepackage[utf8]{inputenc}

\newtcbox{\blue}[1][]{on line,boxsep=2pt,left=0pt,right=0pt,top=0pt,bottom=0pt,colframe=white,colback=blue!30!white,#1}
\newtcbox{\red}[1][]{on line,boxsep=2pt,left=0pt,right=0pt,top=0pt,bottom=0pt,colframe=white,colback=red!30!white,#1}

\title{Rethinking Relation Classification with Graph Meaning Representations}

\newcommand{\uestc}{$^1$}
\newcommand{\ku}{$^2$}

\author{Li Zhou\uestc \ku, Wenyu Chen\uestc, Dingyi Zeng\uestc, Malu Zhang\uestc, Daniel Hershcovich\ku \\
{\uestc}University of Electronic Science and Technology of China\\
{\ku}Department of Computer Science, University of Copenhagen \\
\texttt{\small (li\_zhou, zengdingyi)@std.uestc.edu.cn, (cwy, maluzhang)@uestc.edu.cn, dh@di.ku.dk}
}

\begin{document}
\maketitle
\begin{abstract}
In the field of natural language understanding, the intersection of neural models and graph meaning representations (GMRs) remains a compelling area of research. 
Despite the growing interest, a critical gap persists in understanding the exact influence of GMRs, particularly concerning relation extraction tasks. 
Addressing this, we introduce DAGNN-plus, a simple and parameter-efficient neural architecture designed to decouple contextual representation learning from structural information propagation. 
Coupled with various sequence encoders and GMRs, this architecture provides a foundation for systematic experimentation on two English and two Chinese datasets. 
Our empirical analysis utilizes four different graph formalisms and nine parsers. 
The results yield a nuanced understanding of GMRs, showing improvements in three out of the four datasets, particularly favoring English over Chinese due to highly accurate parsers. 
Interestingly, GMRs appear less effective in literary-domain datasets compared to general-domain datasets. These findings lay the groundwork for better-informed design of GMRs and parsers to improve relation classification, which is expected to tangibly impact the future trajectory of natural language understanding research.

\end{abstract}

\section{Introduction}
\begin{figure}[t]
    \centering
    \includegraphics[width=.9\linewidth]{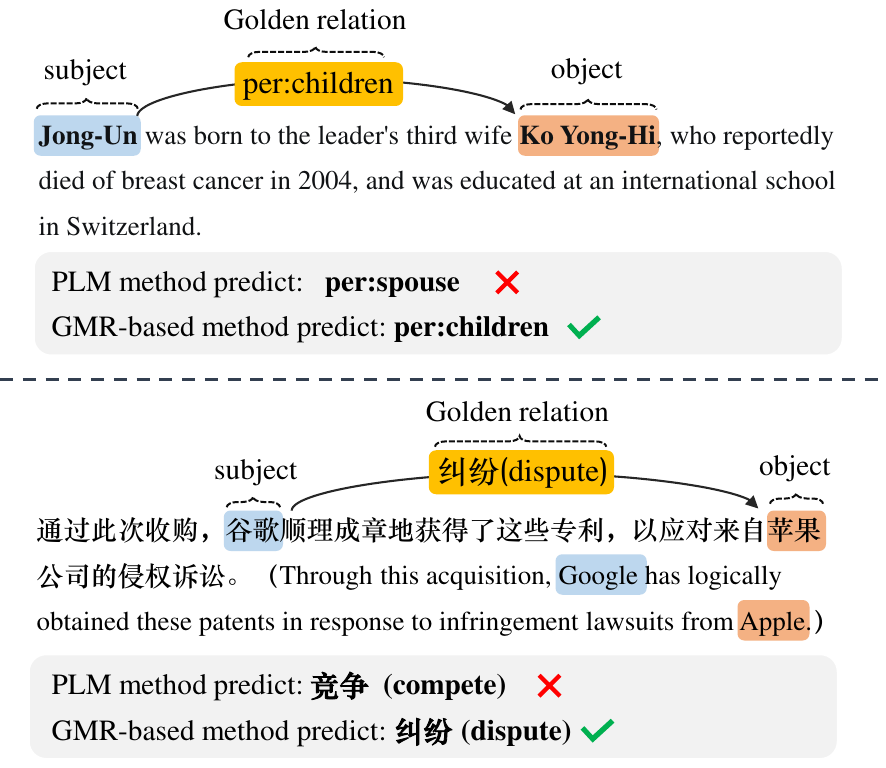}
    \caption{Example instances for the RC task from the RC datasets we experiment with (\S\ref{sec:datasets}). The English sample (top) is selected from the Re-TACRED dataset \cite{stoica2021re}, and the Chinese sample (bottom, with English translations in parentheses) is selected from the FinRE dataset \cite{li-etal-2019-chinese}. 
    % The neurosymbolic method uses DM formalism (\S\ref{sec:frameworks}) parsed by AM-parser (\S\ref{sec:parsers}) and GCN-res graph encoder (\S\ref{sec:encoders}) for English, and the syntactic dependencies parsed by DDparser and DAGNN-plus graph encoder for Chinese.
    }
    \label{fig:RC_example}
\end{figure}

Relation extraction (RE) is an important natural language processing (NLP) task across various downstream tasks and domains, such as biomedical information extraction \cite{papanikolaou-etal-2022-slot} and text mining for knowledge base enrichment \cite{trisedya-etal-2019-neural, LI2022346}. In this paper, we focus on the relation classification (RC) subtask, which aims to classify the relation that holds between a given pair of entities.  Figure~\ref{fig:RC_example} illustrates examples of relation classification (RC), featuring one instance in English and one in Chinese. In the English example, the subject entity is "Jong-Un," the object entity is "Ko Yong-Hi," and their relation within this sentence is denoted as "per: children."

Many neural RC methods have been proposed for learning to capture contextual information in the representation during end-to-end training~\cite{dixit-al-onaizan-2019-span, wang-lu-2020-two, huguet-cabot-navigli-2021-rebel-relation, GENG202224}, most of which are based on pre-trained language models \cite[PLMs;][]{devlin-etal-2019-bert, liu2019roberta}.
% And some works shown that syntactic information exists in the attention head and hidden representation of pre-trained language models \cite[PLMs;][]{devlin-etal-2019-bert, liu2019roberta}~\cite{tenney-etal-2019-bert, clark-etal-2019-bert, ravishankar-etal-2021-attention}. 
However, PLMs without explicit structural information still face limitations when it comes to RC tasks~\cite{tian-etal-2022-improving}.
On the one hand, words that have strong associations with specific relations can potentially lead to misunderstandings in PLMs. This is because PLMs, as neural sequence models, are fast and learn patterns from data~\cite{nye2021improving}. During the fine-tuning process of PLMs, they will learn from the data experience within the training set, which will influence the model's generalization ability. We conduct a statistic regarding the relation "per:spouse" in the training set of ReTACRED~\cite{stoica2021re}. Our findings indicate that 67.5\% of cases contain direct keywords associated with "per:spouse" such as "wife","husband" and "hubby". Additionally, 24.7\% of cases include indirect keywords related to "per:spouse" such as "wedding," "marry," and "fiancé". Only 7.7\% of cases do not feature any of these cue words used for relation prediction. 
So the word ``wife'' in the English case of Figure~\ref{fig:RC_example}, can confuse the PLM method and causes it to predict a wrong relation ``per:spouse''. 
On the other hand, the knowledge encoded in PLMs can also mislead them \cite{sogaard-2021-lockes}. Because there exists belief bias in PLMs~\cite{sogaard-2021-lockes, gonzalez-etal-2021-interaction, lent-sogaard-2021-common}. Belief bias, a form of cognitive bias, is defined in psychology as the systematic (non-logical) inclination to assess a statement based on one's prior beliefs rather than its logical validity~\cite{klauer2000belief, evans1983conflict}. The corpus used for training PLMs is a prior belief, that is, knowledge. 
As illustrated in the Chinese example in Figure~\ref{fig:RC_example}, when assessing the relation between Google and Apple, PLMs may readily categorize it as a competitive one instead of dispute one, due to the well-known rivalry between the two companies.

% So as shown in the Chinese case of Figure~\ref{fig:RC_example}, when evaluating the relation between Google and Apple, it's easy for PLMs to classify it as competition because the fact that Google and Apple are in competition.

Recent works \cite{zhou2020eawgcn, tian-etal-2021-dependency, yellin-abend-2021-paths, zhang-ji-2021-abstract, bassignana2023silver, WAN2023110873} have shown that discrete Graph Meaning Representations (GMRs) are conducive to RC when combined in neural networks. 
Because GMRs can facilitate long-distance word connections among important terms, thereby guiding the system to more effectively extract relation between pairs of entities.
However, most of the current methods employed for GMRs of text, such as GCN~\cite{kipf2017semi} and GAT~\cite{velickovic2018graph}, intertwine representation transformation and propagation operations.
This makes it challenging to determine whether the performance improvement is attributed to the introduction of GMRs or the increase in learning parameters. 
Additionally, these works mainly focus on the syntactic Universal Dependencies (UD) framework \cite{de-marneffe-etal-2021-universal}\footnote{We refer to UD as a GMR since it serves a similar role for our purpose, even though it is often treated as a \textit{syntactic} framework. Generally speaking, syntactic representations, as opposed to semantic ones, are sensitive to some irrelevant syntactic variations and often insensitive to important semantic distinctions \cite{hershcovich-etal-2019-content}.} and on English datasets, overlooking the exploration of different GMRs and the significance of their related parsers' role within them.
    
In this paper, we design a simple and parameter-efficient method to decouple contextual representation learning and structural information propagation for language (\S\ref{sec:encoders}), which can provide a foundation for systematic experiments. Specifically, we use various GMRs, including syntactic and semantic representations (\S\ref{sec:frameworks}), and investigate their related parsers (\S\ref{sec:parsers}). We analyse their contribution to English and Chinese RC (\S\ref{sec:analysisGMR}, \S\ref{sec:analysisparser}). 
% Importantly, the utilization of GMRs relies on parsers for generating graphs from plain text, as obtaining human-annotated graphs is rarely feasible for task-specific data.
% Previous work has focused on methods for using GMRs and neglect the role of parsers. 
% According to our knowledge, we are the first to compare different parsers (\S\ref{sec:parsers}) and investigate whether the quality of parsed GMRs affects the performance in the RC task (\S\ref{sec:analysisparser}). 
% Our experimental pipeline is shown in Figure~\ref{fig:model}.
% To take advantage of structural linguistic information, some approaches directly operate on GMRs by using Graph Neural Networks (GNNs) as encoders, some bias model attention to encourage or force certain tokens to focus on tokens with which they have semantic or syntactic dependencies, and some inject the semantic information into PLMs (\S\ref{sec:related}). However, these previous approaches are complex in design and introduce many parameters. 
% In this paper, we design a simple and parameter-efficient method to decouple contextual representation learning and structural information propagation for language (\S\ref{sec:encoders}).
Specifically, our contributions are:
\begin{itemize}
    % \item We survey different semantic and syntactic GMRs and parsers and their applicability for RC, finding that English parsers are more advanced and widely follow consistent annotation criteria, facilitating evaluation on a shared dataset.
    \item We propose DAGNN-plus, a simple and parameter-efficient method that separates contextual representation learning and structural information propagation for language. This method is compatible with any PLMs and GMRs, offering a robust foundation for systematic experimentation.
    % two simple and lightweight GNN models for encoding GMRs. On two English and two Chinese RC datasets, our models outperform GNN-based and text-only baselines.
    % \item We examine the influence of GMRs and parsers on RC for two English and two Chinese datasets, finding improvements in performance over strong baselines in all datasets except for the Chinese literature one.
    \item We examine the influence of GMRs on RC for two English and two Chinese datasets, finding that incorporating GMRs consistently enhances performance in all datasets, except for the Chinese literature one.
    \item We conduct extrinsic evaluations on different parsers to examine their impact on RC performance. We discover that parsers with good intrinsic evaluations exhibit robustness in downstream RC tasks.
    
    % We analyze the effect of the choice GMR parser on RC performance, finding minor or no sensitivity to parsing performance differences.
    
\end{itemize}

\section{Related Work}\label{sec:related}

% RE: UCCA \cite{yellin-abend-2021-paths}
% IE: AMR \cite{zhang-ji-2021-abstract}
% NLI: DM \cite{wu-etal-2021-infusing}
% MT: AMR \cite{song-etal-2019-semantic, li-flanigan-2022-improving}, UCCA \cite{slobodkin-etal-2022-semantics}
% Text style transfer: UD+SRL \cite{gong-etal-2020-rich}
% Dialogue modeling: AMR \cite{bai-etal-2021-semantic}
% Brain decoding: UD, DM, UCCA \cite{abdou2021does}
% SRL: UD \cite{strubell-etal-2018-linguistically}
% LM: UD/PTG/PTB/EDS \cite{prange-etal-2022-linguistic}
% Semantic textual similarity (STS): UCCA \cite{BOLUCU2022119103}

% \subsection{Neurosymbolic RE}

Recently, there has been growing interest in applying graph meaning representations to NLP models for better representation learning. Existing methods that attempt to inject such structural information are considered from three perspectives: graph encoders, the application of a GNN on the output of a contextualized encoder---referred to as \textit{late fusion} by \citet{sachan-etal-2021-syntax}; and guided attention, which infuses structure into attention layers---referred to as \textit{joint fusion} by \citet{sachan-etal-2021-syntax}. We separate the latter according to whether guided attention is provided during prediction or as training signal.

\subsection{Graph Encoders} With the success of graph neural networks (GNNs) on data involving large graphs, many works have been devoted to exploiting semantic structural information in natural language with GNNs.
Some works~\cite{zhang-etal-2018-graph, zhou2020eawgcn} directly apply Graph Convolution Network \cite[GCN;][]{kipf2017semi} to pruned UD trees for RC task. \citet{yellin-abend-2021-paths} apply GCN to another GMR framework, namely UCCA \cite{abend-rappoport-2013-ucca}, which is reduced into a bi-lexical structure using the conversion algorithm by \citet{hershcovich-etal-2017-transition}. 
% \citet{zhang-etal-2018-graph} design an architecture called C-GCN (Contextualized Graph Convolution Network), applying a GCN \cite{kipf2017semi} to pruned UD trees with a Bi-LSTM contextualized layer.
% \citet{zhou2020eawgcn} extend the GCN model to a parameter-free weighted GCN by constructing a logical adjacency matrix, broadening the information-dependent receptive field. 
% On the other hand, \citet{yellin-abend-2021-paths} apply the C-GCN framework to broad coverage semantic structures for RE---they reduce another GMR framework, namely UCCA \cite{abend-rappoport-2013-ucca}, into a bi-lexical structure using the conversion algorithm by \citet{hershcovich-etal-2017-transition}. 
Heterogeneous Graph Transformer (HetGT) and its variant are proposed to encode AMR graphs \cite{banarescu-etal-2013-abstract} for improving generation tasks~\cite{yao-etal-2020-heterogeneous, li-flanigan-2022-improving}.
% \citet{yao-etal-2020-heterogeneous} propose the Heterogeneous Graph Transformer (HetGT), which adaptively models the various relations in different representation subgraphs of the original graph. 
% \citet{li-flanigan-2022-improving} augments the Transformer model with HetGT, encoding source sentences with AMR graphs \cite{banarescu-etal-2013-abstract} for improving neural machine translation.
\citet{wu-etal-2021-infusing} use R-GCN~\cite{schlichtkrull2018modeling} to encode DM structures, a bi-lexical GMRs \cite{ivanova-etal-2012-contrastive}.
% \citet{zhang-ji-2021-abstract} introduce an attention-based semantic graph aggregator and apply it to AMR graphs to capture global inter-dependency between candidate nodes.
\citet{sachan-etal-2021-syntax} design a syntax-based graph neural network (syntax-GNN) using UD, a variation of the transformer encoder where the self-attention sublayer is replaced by graph attention network \cite[GAT;][]{velickovic2018graph}, then try \textit{late fusion} and \textit{joint fusion} for RC task simultaneously. The authors observe that the \textit{late fusion} model obtains improvement while the \textit{joint fusion} model leads to a drop of F1 points in performance. \citet{10138896} propose a semantic and syntactic aware GAT for the multi-label classification problem, and state that semantic representation outperforms syntactic representation in emotion classification.
% \footnote{Our experimental framework adopt \textit{joint fusion} to explore on RC, and decouple contextual representation learning and structural information propagation.}
 
% The first approach involves a sequential assembly of a transformer and a syntax-GNN , while the second approach interleaves syntax-GNN embeddings within transformer layers. 
% feed BERT contextual representations \cite{devlin-etal-2019-bert} to a syntax-GNN encoder using UD and apply a Highway Gate at its output to adaptively select useful representations. 

\subsection{Guided Attention During Prediction}
Transformers \cite{vaswani2017attention} excel at managing long-term sequential dependencies by using self-attention to represent each word based on information from other words in the sequence.
% Transformers \cite{vaswani2017attention} are successful in handling long-term sequential dependencies. Their self-attention mechanism consists of learning to represent each word by combining feature information from other words in a sequence. 
This can be viewed as a case of a GNN applied on a fully connected graph of words \cite{joshi2020transformers}. However, this architecture does not leverage the graph structure inductive bias \cite{dwivedi2021generalization}.
To explicitly address this, some approaches focus on designing a semantically guided attention layer.
\citet{bugliarello-okazaki-2020-enhancing} propose a parameter-free local self-attention mechanism that enables the model to prioritize the dependency parent of each token during source sentence encoding.
\citet{zhang2020sg} introduce a syntax-guided self-attention layer that limits attention to a word and its ancestor head words. Additionally, \citet{li-etal-2021-improving-bert} propose syntax-aware local attention, which constrains attention scopes based on distances in the syntactic structure.
% \citet{wang-etal-2019-self} integrate absolute and relative structural position embedding into the self-attention mechanism, which is complementary to the standard sequential positional representations.
All the above works are based on UD structure, where the nodes in the UD tree are associated with the words in the sentence. 
Conversely, \citet{slobodkin-etal-2022-semantics} use the UCCA graph of a sentence to generate a scene-aware mask for the self-attention heads of the Transformer encoder.

\subsection{Guided Attention During Training.}
Some approaches inject GMRs into models during the training process as supervision.
\citet{strubell-etal-2018-linguistically} present a multi-task neural model to incorporate auxiliary syntactic information, which can restrict each token to attend to its syntactic parent in one extra attention head.
\citet{xu-etal-2021-syntax} introduce a syntax-aware pre-training task, named dependency distance prediction task (DP), to encourage models to capture global syntactic relations among tokens.
\citet{abdou2021does} employ a fine-tuning method utilizing structurally guided attention for injecting structural bias into language model representations.
% \citet{wang-etal-2021-k} pre-train a knowledge-specific adapter on the task of dependency relation prediction, which aims to predict the head index of each token in the given sentence.
\citet{bai-etal-2022-graph} investigate graph self-supervised training to improve the structure awareness of PLMs over AMR graphs.

\begin{figure}[t]
\centering
\includegraphics[width=0.8\linewidth]{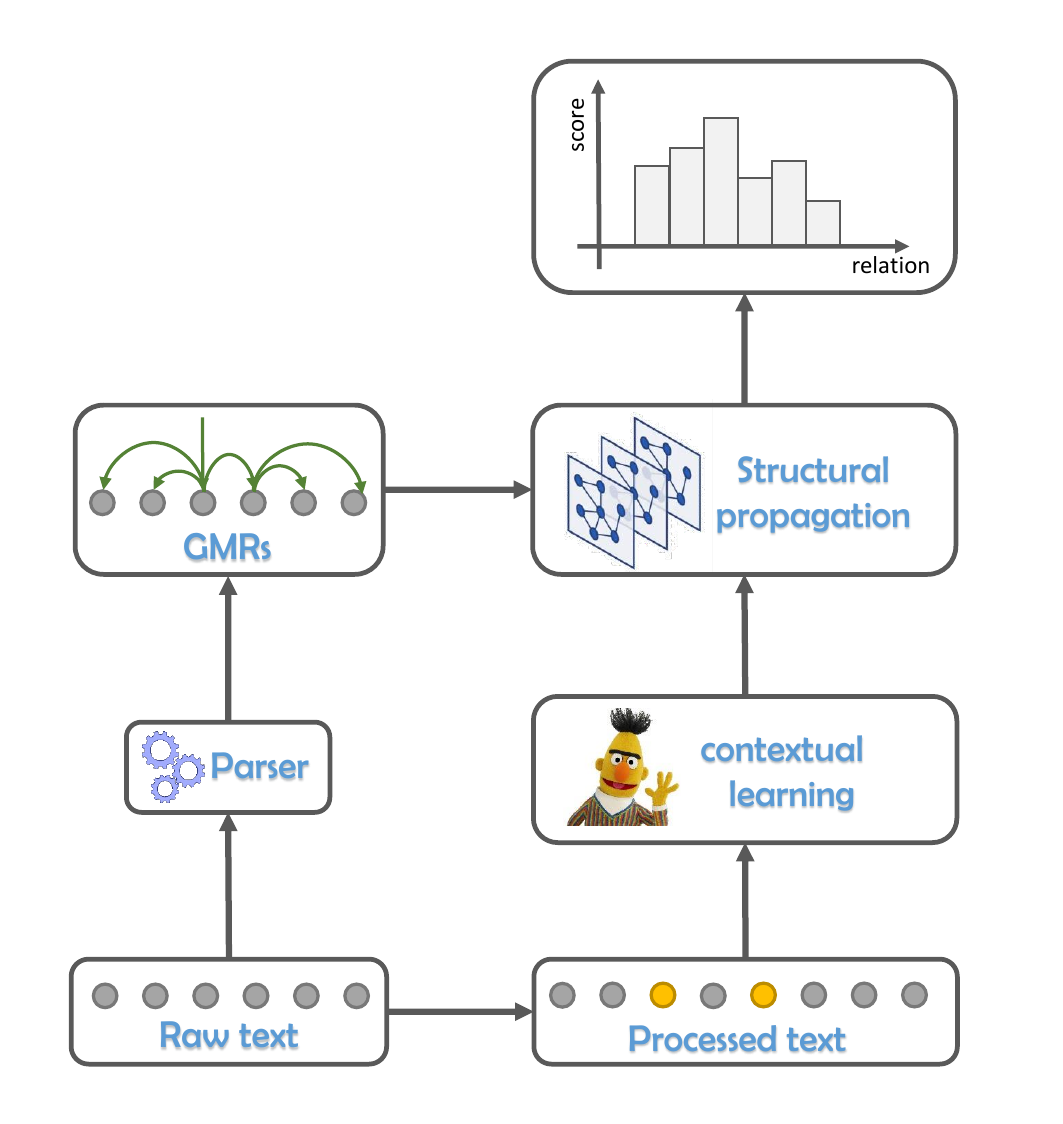}
\caption{The experimental pipeline configuration for the \textit{relation classification} task.
% where text is encoded by a \textit{graph encoder} that takes as input both a \textit{sequence encoder} and a \textit{meaning representation} graph produced by a \textit{parser}. We experiment with different sequence encoders, parsers, meaning representations and graph encoders on four relation classification datasets in two languages.
}
\label{fig:model}
\end{figure}

\begin{figure*}[ht]
    \centering
    \includegraphics[width=1.0\linewidth]{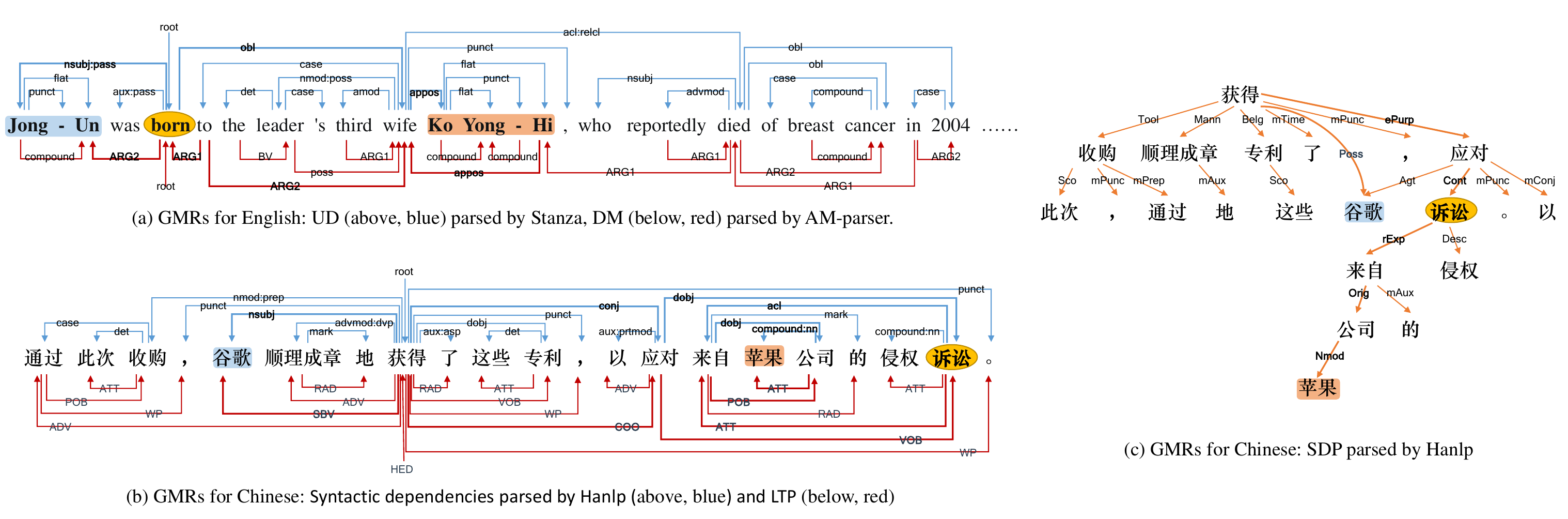}
    \caption{Example GMR graphs for the same two cases shown in Figure~\ref{fig:RC_example}. UD, DEP and SDP attach all words, while DM only connects content words. Each word in UD and DEP only has a single head, and some words in DM and SDP have multiple heads, potentially leading to undirected cycles. All the GMRs can provide long-distance word connections, and offer clue words (\textbf{oval mark}) appearing on the shortest path (\textbf{bold line}) of an entity pair.}
    \label{fig:GMR}
\end{figure*}

\section{Relation Classification with PLMs}
We conduct our experiments based on an approach referred to as ``an improved RC baseline'' \cite[IRC;][]{zhou-chen-2022-improved} which is proposed to predict the relation between the given subject entity and object entity in a sentence, out of a set $\mathcal{R}$ of possible relations or $\mathrm{NA}$ for no relation.

Given the input text $\bm{x}$, IRC firstly marks the entity spans with special tokens in the start and end position (see Figure~\ref{fig:edge_type} for an example).\footnote{In this paper, we utilize two methods employing special tokens, namely Entity Marker and Type Entity Marker, depending on whether the dataset includes entity types.}
In this way, both the subject and object entities span several tokens in the sequence, including text tokens and special tokens.
The processed text $\left\{ w_0, w_1, …, w_{n-1} \right\}$ is fed into a PLM, obtaining the hidden state $\boldsymbol{h}_{i}$ of each token from the last layer.
IRC takes the hidden states of the first token in the span of the subject and object entities as subject embedding $\boldsymbol{h}_{\mathrm{subj}}$ and object embedding $\boldsymbol{h}_{\mathrm{obj}}$. Finally, $\boldsymbol{h}_{\mathrm{subj}}$ and $\boldsymbol{h}_{\mathrm{obj}}$ are concatenated, projected by $W_\mathrm{proj}$ and RELU activation, and fed into a softmax classifier parameterized by $W_r$ and $b_r$ for each relation:
\begin{equation}
\begin{aligned}
\boldsymbol{z}&=\mathrm{RELU}\left( \boldsymbol{W}_{\mathrm{proj}}\left[ \boldsymbol{h}_{\mathrm{subj}}, \boldsymbol{h}_{\mathrm{obj}} \right] \right), \\
P\left( r|\bm{x} \right) &=\frac{\exp \left( \boldsymbol{W}_r\boldsymbol{z}+\boldsymbol{b}_r \right)}{\sum_{\hat{r}\in \mathcal{R} \cup \left\{ \mathrm{NA} \right\}}{\exp \left( \boldsymbol{W}_{\hat{r}}\boldsymbol{z}+\boldsymbol{b}_{\hat{r}} \right)}}.
\end{aligned}
\label{eq:softmax_classifier}
\end{equation}

We design our RC exploration framework with GMRs in \S\ref{sec:methods} based on the IRC method~\cite{zhou-chen-2022-improved}, which outperforms previous models that have a more complex architecture.

\section{Encoding Meaning Representations with Graph Neural Networks}\label{sec:methods}
Figure~\ref{fig:model} shows our experimental framework, which combines GMRs with PLMs for RC.
Firstly, following \citet{zhou-chen-2022-improved}, we add special tokens for entity spans and adopt PLMs as sequence encoders for obtaining contextual representations.
Then we parse the original text into GMRs and design a simple and parameter-efficient graph encoder for extracting syntactic/semantic structural information from the parser outputs.
Finally, the softmax classifier (Eq.~\ref{eq:softmax_classifier}) is applied, with the final encoded representations as input, to predict the relation.

\subsection{GMR Frameworks}\label{sec:frameworks}
To explore the contribution of GMRs, we experiment with distinct formalisms for representation of syntactic/semantic structure, which come from different linguistic traditions and can capture different aspects of linguistic signal. The example graph for each formalism is shown in Figure~\ref{fig:GMR}.
In the English case, the paths between the subject and object in GMRs include the keyword ``born'', which serves as an effective clue for correctly inferring the relation `per:children'. Similarly, in the Chinese case, the keyword for relation inference is ``\begin{CJK}{UTF8}{gbsn}诉讼\end{CJK}'' (lawsuits).

\paragraph{Syntactic representations.}
For syntactic representation in English, we select \textbf{UD} \cite[Universal Dependencies;][]{nivre-etal-2020-universal}, a popular multilingual framework for annotating lexical dependencies, in which the syntactic structure of a sentence is represented by the relationships between its words.
For Chinese, various standards exist for syntactic dependency annotation, including UD, Chinese Stanford Dependencies \cite{chang-etal-2009-discriminative}, Penn Chinese Treebank \cite{xue2005penn} and Baidu Chinese Treebank \cite{Qiang2004AnnotationSF}. Each parser (\S\ref{sec:parsers}) uses its own standard, often without specifying the version or training set. Therefore, for Chinese syntactic dependencies, we are mainly interested in comparison between the parsers rather than GMR framework. For convenience of description, we refer to the Chinese syntactic dependency frameworks of these parsers collectively as \textbf{DEP}.

\paragraph{Semantic representations.}
For semantic representations, we select \textbf{DM} \cite[DELPH-IN MRS Bi-Lexical Dependencies;][]{ivanova-etal-2012-contrastive} for English and \textbf{SDP} \cite[Chinese Semantic Dependency Parsing;][]{che-etal-2016-semeval} for Chinese. DM is based on MRS \cite[Minimal Recursion Semantics;][]{copestake2005minimal}, and represents the meaning of a sentence as a graph of interconnected nodes. 
SDP extends the traditional tree-structured syntactic representation of Chinese sentences to directed acyclic graphs that can capture richer latent semantics.
% SDP extends traditional tree-structured representation of Chinese sentences to directed acyclic graphs that can capture richer latent semantics. 

\paragraph{Synthetic vs Semantic}  Semantic representations are concerned with the meaning of words and the overarching message conveyed by a sentence. In contrast, syntactic representations are concerned with the grammatical arrangement of words within the sentence. In the example ``She reads a book'' the semantic structure clarifies that the sentence signifies a woman engaging in the act of reading a book. On the other hand, the syntactic structure illustrates how words are organized grammatically in the sentence, adhering to the Subject-Verb-Object (SVO) structure mandated by English grammar.

\begin{table}[t]
\centering
\small
\begin{tabular}{l|l}
\toprule
\textbf{English} & \textbf{Chinese} \\
\midrule
\textbf{UD} \cite{nivre-etal-2020-universal}                            & \textbf{DEP} (various standards) \\ \hline
Stanza \cite{qi-etal-2020-stanza}                                       & Stanza \\
UDPipe1 \cite{straka-etal-2016-udpipe}                                  & Hanlp \cite{he-choi-2021-stem} \\
UDPipe2 \cite{straka-2018-udpipe}                                       & LTP \cite{che-etal-2021-n} \\
                                                                        & \multirow{2}{3cm}{DDparser \cite{zhang2020practical}} \\\\
\midrule
\textbf{DM} \cite{{ivanova-etal-2012-contrastive}}                      & \textbf{SDP} \cite{che-etal-2016-semeval} \\ \hline
ACE \cite{wang-etal-2021-automated}                                     & LTP \\
\multirow{2}{3cm}{AM-parser \cite{lindemann-etal-2019-compositional}}   & Hanlp \\\\
HIT-SCIR \cite{che-etal-2019-hit}                                       & \\
\bottomrule
\end{tabular}
\caption{Parsers for each language and GMR framework. Top: syntactic parsers. Bottom: semantic parsers.}\label{tab:parsers}
\end{table}

\begin{table}[t]
\centering
\setlength{\tabcolsep}{5pt}
\scalebox{0.8}{
\begin{tabular}{l|lll|lll}
\toprule
\textbf{GMR}        & \multicolumn{3}{c|}{\textbf{UD}}    & \multicolumn{3}{c}{\textbf{DM}}  \\ \midrule
\small Benchmark & \multicolumn{3}{c|}{\small CoNLL 2018 (UAS/LAS)} & \multicolumn{3}{c}{\small CoNLL 2019 (SDP F1)}  \\
\small Parser    & \scriptsize Stanza & \scriptsize UDPipe1 & \scriptsize UDPipe2 & \scriptsize ACE      & \scriptsize AM       & \scriptsize HIT-SCIR               \\ \midrule
\small Unlabeled & 86.22  & 80.22   & 85.01   & -    & -   & -       \\
\small Labeled & 83.59  & 77.03   & 82.51 &  93.04$^{\dagger}$   & 94.70  & 95.08      \\ \bottomrule
\end{tabular}}
\caption{Parsing performance on English benchmarks for each parser, CoNLL 2018 UD Shared Task \cite{zeman-etal-2018-conll} and CoNLL 2019 MRP Shared task \cite{oepen-etal-2019-mrp}. ${\dagger}$ marks results from our implementation.}\label{tab:parser_performance}
\end{table}

\subsection{Parsers}\label{sec:parsers}

% \begin{table*}[t]
% \scalebox{0.9}{
% \begin{tabular}{l|llll|lll|ll|ll}
% \toprule
% \textbf{MRs}        & \multicolumn{4}{c|}{\textbf{UD for En}}    & \multicolumn{3}{c|}{\textbf{DM for En}} & \multicolumn{2}{c|}{\textbf{UD for Zh}} & \multicolumn{2}{c}{\textbf{SDP for Zh}} \\ \midrule
% Benchmark & \multicolumn{4}{c|}{CoNLL 2018 Shared Task (English-EWT)} & &  \\ \hline
% Parsers    & Sf. & Sz. & pipe1. & pipe2. & ACE      & AM       & HIT.     & Sz. & Han.  & LTP     & Han.             \\ \midrule

% UAS | F1 | UAS | F1 & 86.22    & 86.22  & 80.22   & 85.01   &  ${93.04}^{\dagger}$   & 94.70  & 95:08   & ${74.97}^{\dagger}$    &  ${51.45}^{\dagger}$    & 80.4   & ${0.00}^{\dagger}$   \\
% LAS | -\quad | LAS | - & 83.69    & 83.59  & 77.03   & 82.51 &  -    & -   & -  & ${71.82}^{\dagger}$    &   ${45.71}^{\dagger}$     &  -    &  -      \\ \bottomrule
% \end{tabular}}
% \caption{The evaluation results on related benchmark for each parsers. ${\dagger}$ marks results from our implementation.}
% \label{tab:parser_performance}
% \end{table*}

% \footnote{To evaluate UD parsing performance of Hanlp on the same benchmark, we adopt a cross-linguistic  parsing model, whose annotation standard is similar with the benchmark. But when parsing for Chinese dataset, we adopt Chinese parsing model, whose annotation standard Stanford Dependencies Chinese.}

To investigate whether the quality of GMRs obtained by their related parsers affects the performance in the RC task, we generate GMRs structure for each formalism using multiple parsers. 
Table~\ref{tab:parsers} lists the parsers we use for each language and GMR framework.
We report the parser performance in relevant English shared tasks in Table~\ref{tab:parser_performance}.\footnote{Chinese parsers are not evaluated here because of lacking uniform benchmark results, different annotation standards for the same formalism and non-freely-available evaluation data.}
We evaluate UD parsers on the English-EWT test set of the CoNLL 2018 Shared Task \cite{zeman-etal-2018-conll}, in which the metrics \textit{UAS} and \textit{LAS} are reported.
We evaluate DM parsers on CoNLL 2019 MRP shared task \cite{oepen-etal-2019-mrp}, with the metric SDP F1 score.
% for Chinese on CoNLL 2018 Shared Task (Chinese-GSDSimp)\footnote{\url{https://github.com/UniversalDependencies/UD\_Chinese-GSDSimp}}, 
% SDP parsers on SemEval 2016 Task 9 \cite{che-etal-2016-semeval}\footnote{DEP parsers for Chinese are not evaluated because of the missing results from the same benchmark and the paid evaluation data.},

Since the parsers we use are all near state-of-the-art (SOTA), we observe small differences in performance for both frameworks. However, non-trivial differences do exist. This will allow us to analyze the relation between parser performance and contribution to RC in \S\ref{sec:analysisparser}.

\begin{figure}[t]
    \centering
    \includegraphics[width=.9\linewidth]{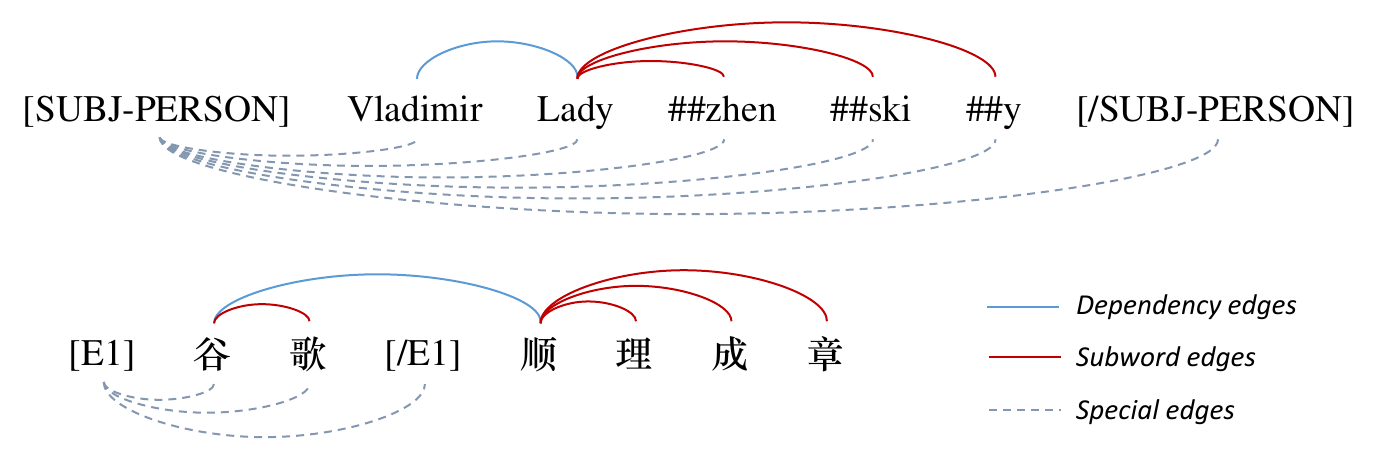}
    \caption{An example diagram of three edge constructs. \textit{Dependency edges} connect the first subwords of words connected in the original GMR, \textit{subword edges} connect all subwords of a word to the first subword and \textit{special edges} connect all subwords of an entity, as well as the closing special token, to the opening special token.}
    \label{fig:edge_type}
\end{figure}

\subsection{Graph Encoder}\label{sec:encoders}
GCN \cite{kipf2017semi} and GAT \cite{velickovic2018graph} are popular graph encoders in NLP. However, their application to underlying PLM representations can lead to the loss of learned contextual information due to the over-smoothing problem \cite{chen2020measuring}. Additionally, the introduction of GCN and GAT increases the number of trainable parameters. To address these issues, we propose DAGNN-plus, a simple and parameter-efficient method based on DAGNN \cite{liu2020towards}. DAGNN-plus combines learned textual representations with the inductive bias from GMRs.

% GCN \cite{kipf2017semi} and GAT \cite{velickovic2018graph} are the most popular graph encoders in NLP. However, because of the over-smoothing problem \cite{chen2020measuring}, learned contextual information can be easily lost by applying these GNNs to underlying representations generated by PLMs. Besides, the introduction of GCN and GAT brings about an increase in the number of trainable parameters.
% Therefore, we design a simple and parameter-efficient method DAGNN-plus, which is based on DAGNN \cite{liu2020towards}.
% We apply DAGNN-plus to combine the learned textual representations with the inductive bias obtained from GMRs.

\paragraph{Graph encoders over subwords}
PLMs operate on subword units, often using BPE \cite{sennrich-etal-2016-neural} or WordPiece \cite{wu2016google} instead of linguistic tokens (words). We transform the original dependency edges between words into the \textit{dependency edges} between the first subword of head words and the first subword of tail words. We follow \citet{sachan-etal-2021-syntax} in applying \textit{subword edges} from the first (head) subword of a token to the remaining (tail) subwords of the same token. Additionally, we incorporate novel \textit{special edges} because of the introduction of the special tokens that indicate the entity spans \cite{zhou-chen-2022-improved}. In the experiments, we ignore the direction and type of all edges, which makes the input of graph structure very simple. An example diagram of these three edge constructs is in Figure~\ref{fig:edge_type}.

% \paragraph{GCN-res.} We add residual connections to the output in each GCN layer, defined as follows:
% \[
%     h_i^{(l+1)} = \sigma\left(\sum_{j \in N_i} \frac{1}{|N_i|} W^{(l)} h_j^{(l)} + b^{(l)}\right) + h_i^{(0)},
% \]
% where $h_i^{(l)}$ is the representation of node $i$ at layer $l$, $h_i^{(0)}=x_i$, $N_i$ is the set of neighbors of node $i$, $|N_i|$ is the number of neighbors, $W^{(l)} \in R^{d \times d}$, $b^{(l)} \in R^{d}$ are learnable parameters at layer $l$,  and $\sigma(\cdot)$ is a non-linear activation function, such as a rectified linear unit (ReLU). The number of the newly introduced parameters in GCN-res are $L \times (d \times d + d)$, where $L$ is the number of layers and $d$ is the layer dimension. 

\paragraph{DAGNN-plus} 
In vanilla GCN, the entanglement of representation transformation and propagation in each layer leads to an increased parameter count and a potential loss of discernible information. This poses a challenge in determining the primary factor influencing downstream performance differences between representation transformation and information propagation. DAGNN \cite{liu2020towards} decouples the two operations, which can alleviate the over-smoothing problem. Inspired by this, we introduce DAGNN-plus, a framework capable of incorporating any PLMs for capturing contextual information, as well as any GMRs that provides structural bias. The representation transformation operation for contextual learning is defined using PLMs, which serves as sequence encoders, as follows:
\begin{align}
\left( \boldsymbol{h}_0, \boldsymbol{h}_1, …, \boldsymbol{h}_{n-1} \right) =\mathrm{PLMs}\left( w_0, w_1, …, w_{n-1} \right) 
\end{align}
where $\boldsymbol{h}_i$ denotes the hidden representation of the token $w_i$ generated by PLMs. The information propagation operation for structural propagation, served as graph encoders, is defined as following:
\begin{align}
\boldsymbol{q}_{i}^{(l+1)}=\sigma \left( \sum_{j\in N_i}{\frac{1}{|N_i|}}\boldsymbol{q}_{j}^{(l)} \right) &\in R^d,\label{eq:rep_propb}\\
\tilde{\boldsymbol{q}}_i=\mathrm{stack}\left( \boldsymbol{h}_i, \boldsymbol{q}_{i}^{(1)}, …, \boldsymbol{q}_{i}^{(L)} \right) & \in R^{\left( L+1 \right) \times d},\\
\boldsymbol{s}_i=\mathrm{sigmoid}\left( \tilde{\boldsymbol{q}}_i\boldsymbol{W} \right) & \in R^{\left( L+1 \right) \times 1},\\
\boldsymbol{h}_{i}^{out}=\mathrm{squeeze}\left( \boldsymbol{s}_{i}^{\mathrm{T}}\tilde{\boldsymbol{q}}_i \right)& \in R^d,
\label{eq:rep_propa}
\end{align}
% where Eq.~\ref{eq:rep_trans} represents the representation transformation operation, and Eq.~\ref{eq:rep_propb}-\ref{eq:rep_propa} the representation propagation operation.
where $\boldsymbol{q}_{i}^{(0)}=\boldsymbol{h}_i$ denotes the initial node representation, $N_i$ represents the neighboring token of token $w_i$ in the given GMR structure.
$\boldsymbol{q}_{i}^{(l)}$ represents the representation after the $l_{th}$ round of structural information propagation. Specially, $\boldsymbol{W} \in R^{d \times 1}$ is the only learnable parameter introduced in structural information propagation, which is employed to learn the receptiveness $\boldsymbol{s}_i$ across each layer of structural information. After obtaining the enriched representation $\boldsymbol{h}_{i}^{out}$, we use IRC method (Eq.~\ref{eq:softmax_classifier}) for downstream RC tasks.
% In our DAGNN-plus,  we retain the representation propagation operation (Eq.~\ref{eq:rep_propb}-\ref{eq:rep_propa}), while replacing the original MLP network, which serves as the representation transformation operation (Eq.~\ref{eq:rep_trans}), with the PLM sequence encoder itself.
In our DAGNN-plus, when combined with GMRs, the number of parameters only increases by $d \times 1 = d$ at any setting compared with raw baseline IRC, making it a highly parameter-efficient approach.
% resulting in a constant time complexity of $O(d)$.
% which can combine with GMRs, only $d \times 1 = d$ parameters compared with PLMs, leading to a constant time complexity of of $O(d)$. 
By this way, we can separate contextual representation learning and structural information propagation for language text, and ensure that the performance difference is solely attributed to the information propagation based on GMR.

\begin{table}[t]
\centering
\scalebox{0.9}{
\begin{tabular}{lrrrr}
\toprule
\multicolumn{1}{l}{\multirow{2}{*}{\textbf{Datasets}}} & \multicolumn{3}{c}{\textbf{Instances}}        & \multicolumn{1}{r}{\multirow{2}{*}{\textbf{Rel.}}} \\ \cline{2-4}
\multicolumn{1}{c}{}     & \multicolumn{1}{r}{\textbf{Train}} & \multicolumn{1}{r}{\textbf{Dev}} & \multicolumn{1}{r}{\textbf{Test}} & \multicolumn{1}{r}{}       \\ 
\midrule
Re-TACRED & 58465 & 19584 & 13418 & 40 \\
SemEval & 8000 & - & 2717 & 19 \\
\midrule
FinRE & 13485 & 1487 & 3727  & 44 \\
SanWen &  17182 & 1792 & 2219  & 10 \\ % include unknow

\bottomrule
\end{tabular}}
\caption{\label{tab:datasets_statistics}
Datasets statistics for RE. The top two datasets are in English and the bottom two are in Chinese. \textbf{Rel.} is the number of unique relations in a dataset.}
\end{table}

\begin{table*}[ht]
    \centering
    \scalebox{0.65}{
    \begin{tabular}{c|m{16cm}|m{5cm}}
    \toprule
    Dataset & Cases & Relation \\
    \midrule
     \multirow{3}{*}{Re-TACRED}    & [1] Americans have a right to know the truth - Islam is a religion of intolerance and violence," said Richard Thompson, legal director of the \blue{Thomas More Law Center} in \red{Ann Arbor}. &  org:city\_of\_branch \\  \cline{2-3}
     
                                & [2] \blue{He} was also president of the \red{Pakistan Boxing Federation} for 33 years, until 2008. & per:employee\_of \\ \cline{2-3}
                                & [3] \blue{She} is pregnant and said their daughter will grow up knowing \red{her} father was an amazing person.& per:children\\
                                % & He is survived by his third wife, former television news correspondent \red{Marilyn Berger}; his sons, Steven and Jeffrey; his daughter, Lisa Cassara; \blue{his} stepdaughter, Jilian Childers Hewitt, whom Hewitt adopted; and three grandchildren. &  per:spouse \\ \cline{2-3}
                                % & Topaz, one of Israel's most famous television stars, had been in jail for several months since the start of \blue{his} trial for allegedly hiring thugs to assault top \red{Israeli} media executives he blamed for keeping him off the air. & per:origin \\ \cline{2-3}
                                % & At the same time, Chief Financial Officer \blue{Douglas Flint} will become \red{chairman} , succeeding Stephen Green who is leaving to take a government job. &  per:title \\ \cline{2-3}
\midrule
    \multirow{2}{*}{SemEval}                 & [4] Avian \blue{influenza} is an infectious disease of birds caused by type A strains of the influenza \red{virus}. & Cause-Effect (virus, influenza) \\ \cline{2-3}
                            & [5] The \blue{ear} of the African \red{elephant} is significantly larger--measuring 183 cm by 114 cm in the bush elephant.& Component-Whole(ear,elephant) \\
\midrule
    \multirow{1}{*}{FinRE}                 & [6] \begin{CJK}{UTF8}{gbsn}更为巧合的是,\blue{广汇集团}在桂林的子公司——\red{广运实业投资有限责任公司}的执行董事、常务副总经理也叫宋军,也是从新疆来到桂林的。(
Coincidentally, \blue{Guanghui Group}'s subsidiary in Guilin, \red{Guangyun Industrial Investment Co., Ltd.}, has an executive director and deputy general manager named Song Jun, who also came to Guilin from Xinjiang.)  \end{CJK} & \begin{CJK}{UTF8}{gbsn}	拥有\end{CJK} (own) \\ 
\midrule
    \multirow{3}{*}{SanWen}                 & [7] \begin{CJK}{UTF8}{gbsn}\blue{山间}水草丰茂，\red{清泉}潺潺。(Lush greenery and babbling \red{brooks} amid the \blue{mountains})  \end{CJK} & Located \\ \cline{2-3}
    & [8] \begin{CJK}{UTF8}{gbsn} 有\blue{家}回，便产生了\red{故乡}的理念。(When there is a \blue{home}, the concept of \red{hometown} arises.)  \end{CJK} & Part-Whole	 \\ \cline{2-3}
    & [9] \begin{CJK}{UTF8}{gbsn} 今天，\blue{心里乐开了花的爸爸妈妈}见了\red{我}，都有些不好意思地说：“还是女儿有远见。”  (Today, \blue{my parents whose heart is filled with happiness and joy}, saw \red{me} and said somewhat bashfully, "It's our daughter who has foresight.")  \end{CJK} & Family	 \\ 

    \bottomrule
    \end{tabular}}
    \caption{Dataset cases. Blue highlights \blue{Subject Entity} or \blue{Entity 1}, red highlights \red{Object Entity} or \red{Entity 2}. For Chinese cases, the English translations are indicated within parentheses.}
    \label{tab:data_cases}
\end{table*}

\section{Experiments}\label{sec:experiments}
\subsection{Datasets}\label{sec:datasets}

In the RC experiments, we use two English benchmark datasets, \textbf{Re-TACRED} \cite{stoica2021re} and \textbf{SemEval 2010 task 8} \cite[SemEval;][]{hendrickx-etal-2010-semeval}, and two Chinese datasets, \textbf{FinRE} \cite{li-etal-2019-chinese} and \textbf{SanWen} \cite{xu2017discourse}. 
% All datasets are widely used. 
Re-TACRED is a new completely re-annotated version of the TACRED dataset \cite{zhang-etal-2017-position}, in which non-English-written and entity-span-ambiguous sentences are removed.
SemEval is a public dataset with directional relations. The Chinese FinRE dataset is manually annotated from financial news articles in Sina Finance.\footnote{\url{https://finance.sina.com.cn/}} SanWen dataset is constructed from Chinese literature articles. We summary the data statistics of the four datasets in Figure~\ref{tab:datasets_statistics}. Specifically for SemEval, we follow previous studies~\cite{tian-etal-2021-dependency,chen-etal-2021-relation, tian-etal-2022-improving} and use its official train/test split.

Table~\ref{tab:data_cases} shows some cases for each dataset. For the textual expression of four datasets, the cases from the first three datasets are characterized by general expression, conveying information in a concise and straightforward manner, while the last dataset SanWen belongs to literary expression, emphasizing imagery and artistic qualities (case [7] [8]). Besides, the entities in some cases of Re-TACRED and SanWen appear as pronouns (case [2] [3] [9]). Moreover, some entites in SanWen are represented by adjective phrases rather than nouns (case [9]). Therefore, these four datasets differ not only in language but also in the information domain and the form of entity representation.

\subsection{Implementation}
Our models are implemented in PyTorch, based on Hugging Face\footnote{\url{https://huggingface.co/}} for sequence encoders of PLMs and DGL\footnote{\url{https://www.dgl.ai/}} for graph encoders. In sequence encoders, we choose \texttt{BERT}$_{\texttt{base}}$ and \texttt{BERT}$_{\texttt{large}}$ \cite{devlin-etal-2019-bert} for English, \texttt{BERT-base-Chinese} \cite{devlin-etal-2019-bert} and \texttt{Chinese-BERT-wwm} \cite{cui-etal-2020-revisiting, cui-etal-2021-pretrain} for Chinese. During text processing, we adopt \textit{typed entity marker} for Re-TACRED,\footnote{Only Re-TACRED provides subject and object types.} and \textit{entity marker} for SemEval, FinRE and SanWen. For graph encoder, we choose layer $L \in \{2,3,4\}$.
Our models are optimized with the learning rate of $5e-5$ on \texttt{BERT}$_{\texttt{base}}$, \texttt{BERT-base-Chinese} and \texttt{Chinese-BERT-wwm} and $3e-5$ on \texttt{BERT}$_{\texttt{large}}$. We fine-tune the model 5 epochs for English and 10 epochs for Chinese, and run 5 times with different random seeds for each setting.

\subsection{Baselines}
For English datasets, we choose four text-only models: SpanBERT~\cite{joshi-etal-2020-spanbert}, LUKE~\cite{yamada-etal-2020-luke}, SP~\cite{cohensupervised}, and IRC~\cite{zhou-chen-2022-improved}; five GNN-based models: C-GCN \cite{zhang-etal-2018-graph}, EA-WGCN \cite{zhou2020eawgcn}, A-GCN \cite{tian-etal-2021-dependency}, TaMM \cite{chen-etal-2021-relation}, and RE-DMP \cite{tian-etal-2022-improving}.
For Chinese FinRE and SanWen datasets, we reference the existing text-only methods, two of which are BERT variants: SpERT \cite{eberts2019span}, MacBERT \cite{cui-etal-2020-revisiting}, and five of which focus on characters- and words-features for Chinese: LRCNN \cite{ijcai2019p692}, MG Lattice \cite{li-etal-2019-chinese}, SR-BRCNN \cite{10.1007/978-3-030-82147-0_3}, LAN \cite{9559723} and PRM-CNN \cite{zhao2022novel}.

% EnRE
\begin{table}[t]
\centering
\resizebox{\linewidth}{!}{
\begin{tabular}{llrr}
\toprule
\textbf{Model} & \textbf{Encoder}   & \textbf{TAC.} & \textbf{Sem.}  \\
\midrule
\multicolumn{4}{l}{\textbf{Text-only}} \\
SpanBERT \cite{joshi-etal-2020-spanbert}                & BERT$_{\mathrm{large}}$           & 85.30                     & - \\
LUKE \cite{yamada-etal-2020-luke}                       & RoBERTa$_{\mathrm{large}}$        & 90.30                     & -  \\
% \multirow{2}{4cm}{REBEL \cite{huguet-cabot-navigli-2021-rebel-relation}}     & RoBERTa$_{\mathrm{large}}$        & 90.39                     & - \\\\

% LUKE + CoRE \cite{wang-etal-2022-rely}                  & RoBERTa$_{\mathrm{large}}$        & 90.90                     & 88.70 \\ 
% KnowPrompt \cite{10.1145/3485447.3511998}               & RoBERTa$_{\mathrm{large}}$        & \textcolor{blue}{91.30}   & 90.20  \\
SP  \cite{cohensupervised}                              & BERT$_{\mathrm{large}}$           & -                         & 91.90   \\
\hline
\rowcolor{orange!20} IRC \cite{zhou-chen-2022-improved}                      & BERT$_{\mathrm{base}}$            &  87.90                    & 91.22  \\
\rowcolor{gray!20}                                                        & BERT$_{\mathrm{large}}$           &  89.70                    & 91.92 \\
% \rowcolor{gray!20}                                                        & RoBERTa$_{\mathrm{large}}$        &  91.10                    & 92.32 \\
\midrule
\multicolumn{4}{l}{\textbf{GNN-based}} \\
C-GCN \cite{zhang-etal-2018-graph}                      & BiLSTM                            & 80.30                     & 84.80 \\
EA-WGCN \cite{zhou2020eawgcn}                           & BiLSTM                            & 81.04                     & 85.40 \\
A-GCN  \cite{tian-etal-2021-dependency}                 & BERT$_{\mathrm{large}}$           & -                         & 89.85   \\
TaMM \cite{chen-etal-2021-relation}                     & BERT$_{\mathrm{large}}$           & -                         & 90.06  \\
RE-DMP \cite{tian-etal-2022-improving}                  & XLNet$_{\mathrm{large}}$          & -                         & 89.90  \\
\hline
\rowcolor{orange!20}IRC + UD (ours)                                         & BERT$_{\mathrm{base}}$            & 88.38                     & 91.63 \\
\rowcolor{orange!20}                                                                             &                                   & ($\uparrow$ 0.48)                       & ($\uparrow$ 0.41) \\
\rowcolor{gray!20}                                                        & BERT$_{\mathrm{large}}$           & 90.26                     & 92.40 \\
\rowcolor{gray!20}                                                                            &                                   & ($\uparrow$ 0.56)                       & ($\uparrow$ 0.48)  \\
% \rowcolor{gray!20}                                                        & RoBERTa$_{\mathrm{large}}$        & \textcolor{orange}{91.29} & 92.49 \\
\hline
\rowcolor{orange!20}IRC + DM (ours)                                         & BERT$_{\mathrm{base}}$            & 88.69                     & 91.58 \\
\rowcolor{orange!20}                                                                             &                                   & ($\uparrow$ 0.79)                       & ($\uparrow$ 0.36) \\
\rowcolor{gray!20}                                                        & BERT$_{\mathrm{large}}$           & 90.33                     & 92.33  \\
\rowcolor{gray!20}                                                                            &                                   & ($\uparrow$ 0.63)                       & ($\uparrow$ 0.41)  \\
% \rowcolor{gray!20}                                                        & RoBERTa$_{\mathrm{large}}$        & \textbf{\textcolor{red}{91.45}}   & 92.41 \\   

\bottomrule

\end{tabular}}
\caption{Comparison of performance (Test F1) on the English datasets Re-TACRED (TAC.) and SemEval (Sem.). }
\label{tab:overall-results-EnRE}
\end{table}

\begin{table}[t]
\centering
\resizebox{\linewidth}{!}{
\begin{tabular}{llrr}
\toprule
\textbf{Model} & \textbf{Encoder}   & \textbf{FinRE} & \textbf{SanWen}  \\
\midrule
% Lattice LSTM \citeyearpar{zhang-yang-2018-chinese} & 47.41 & 63.88\\
% IF-lattice \citeyearpar{wen-etal-2018-structure} & 50.60 &  68.35\\
\multicolumn{4}{l}{\textbf{Text-only}} \\
SpERT  \cite{eberts2019span}${\dagger}$ & BERT$_{\mathrm{base}}$	&	48.34	&	63.60\\
MacBERT \cite{cui-etal-2020-revisiting}${\dagger}$ & RoBERTa$_{\mathrm{large}}$	&	48.72	&	63.73\\
\hline
LRCNN \cite{ijcai2019p692}${\dagger}$  & CNN & 47.10	& 61.12 \\
MG Lattice \cite{li-etal-2019-chinese} & BiLSTM & 49.26 & 65.61   \\
SR-BRCNN \cite{10.1007/978-3-030-82147-0_3} & Transformer	&	50.60	&	68.35\\
LAN \cite{9559723} & BERT-wwm. 	&	51.35	&	69.85\\
PRM-CNN  \cite{zhao2022novel} & RoBERTa$_{\mathrm{large}}$ 	&	52.97	&	67.72\\
\hline
\rowcolor{orange!20} IRC  \cite{zhou-chen-2022-improved} & BERT-base.  &  57.48 & 76.49  \\
\rowcolor{gray!20} & BERT-wwm. & 59.22 & 77.02 \\
\midrule
\multicolumn{4}{l}{\textbf{GNN-based}} \\
% IRC + UD (ours)     &  BERT-base.           & \textcolor{blue}{58.65}           & 76.95 \\
%                     & BERT-wwm.              & \textbf{\textcolor{red}{59.71}}   & \textcolor{orange}{77.28} \\
% \hline
\rowcolor{orange!20} IRC + DEP (ours)    &  BERT-base.             & 58.65     &  76.87 \\
\rowcolor{orange!20}                    &                           & ($\uparrow 1.17$)          & ($\uparrow 0.38$)           \\
\rowcolor{gray!20}                    & BERT-wwm.                 &  59.71          & 76.85   \\
\rowcolor{gray!20}                    &                           & ($\uparrow 0.49$)          & ($\downarrow 0.17$)           \\
\hline
\rowcolor{orange!20} IRC + SDP (ours) &  BERT-base. & 58.65 & 76.84 \\
\rowcolor{orange!20}                    &                           & ($\uparrow 1.17$)          & ($\uparrow 0.35$)          \\
\rowcolor{gray!20}  & BERT-wwm. & 59.55 & 76.67\\
\rowcolor{gray!20}                    &                           & ($\uparrow 0.33$)          & ($\downarrow 0.35$)           \\

\bottomrule

\end{tabular}}
\caption{Comparison of performance (Test F1) on FinRE and SanWen (Chinese). 
% The bold fonts denote SOTA performance. 
BERT-base. and BERT-wwm. are short for \texttt{BERT-base-Chinese} and \texttt{Chinese-BERT-wwm} separately. ${\dagger}$ marks results from \citet{zhao2022novel}. }
\label{tab:overall-results-ChRE}
\end{table}

\subsection{Experimental Results}
\label{sec:Overall Results}

\paragraph{Overall Results} Table~\ref{tab:overall-results-EnRE} and Table~\ref{tab:overall-results-ChRE} display the results of our experiments on the Test sets of English and Chinese datasets respectively. 
% Detailed implementation information can be found in Appendix~\ref{app:implementation}, while the compared baselines are summarized in Appendix~\ref{app:Baselines}. 
The results reveal performance improvements in three out of four datasets. 
Specifically, 
1) In both English datasets and one Chinese FinRE dataset, the models incorporating GMRs outperform the corresponding IRC baselines. This finding suggests that the explicit integration of GMRs, which capture natural language syntax and semantics, can still directly and effectively enhance language understanding ability, even in the presence of high-performing PLMs. 
2) However, GMRs do not consistently demonstrate their effectiveness in assisting with the SanWen dataset, highlighting the challenges associated with applying GMRs in the context of Chinese literary language. This observation underscores the specific difficulties in leveraging GMRs for tasks involving Chinese literary text.
% Our implementation hyperparameters are shown in Table~\ref{tab:hyperparameter}. 
% Furthermore, we divide the Re-TACRED test set into two groups based on the presence of pronouns within entity mentions and investigate the impact of GMRs on each group. Details of these experiments are provided in Appendix~\ref{app:group_experiments}.
Based on these experimental results, we conduct a more in-depth analysis of the GMR frameworks and their associated parsers using English datasets ReTACRED, SemEval, and Chinese dataset FinRE in \S\ref{sec:analysisGMR}, \S\ref{sec:analysisparser}).

\begin{figure*}[t]
    \centering
    \subfigure[Mention Group]{
    \begin{minipage}[]{0.45\linewidth}
    \centering
    \includegraphics[width=3in]{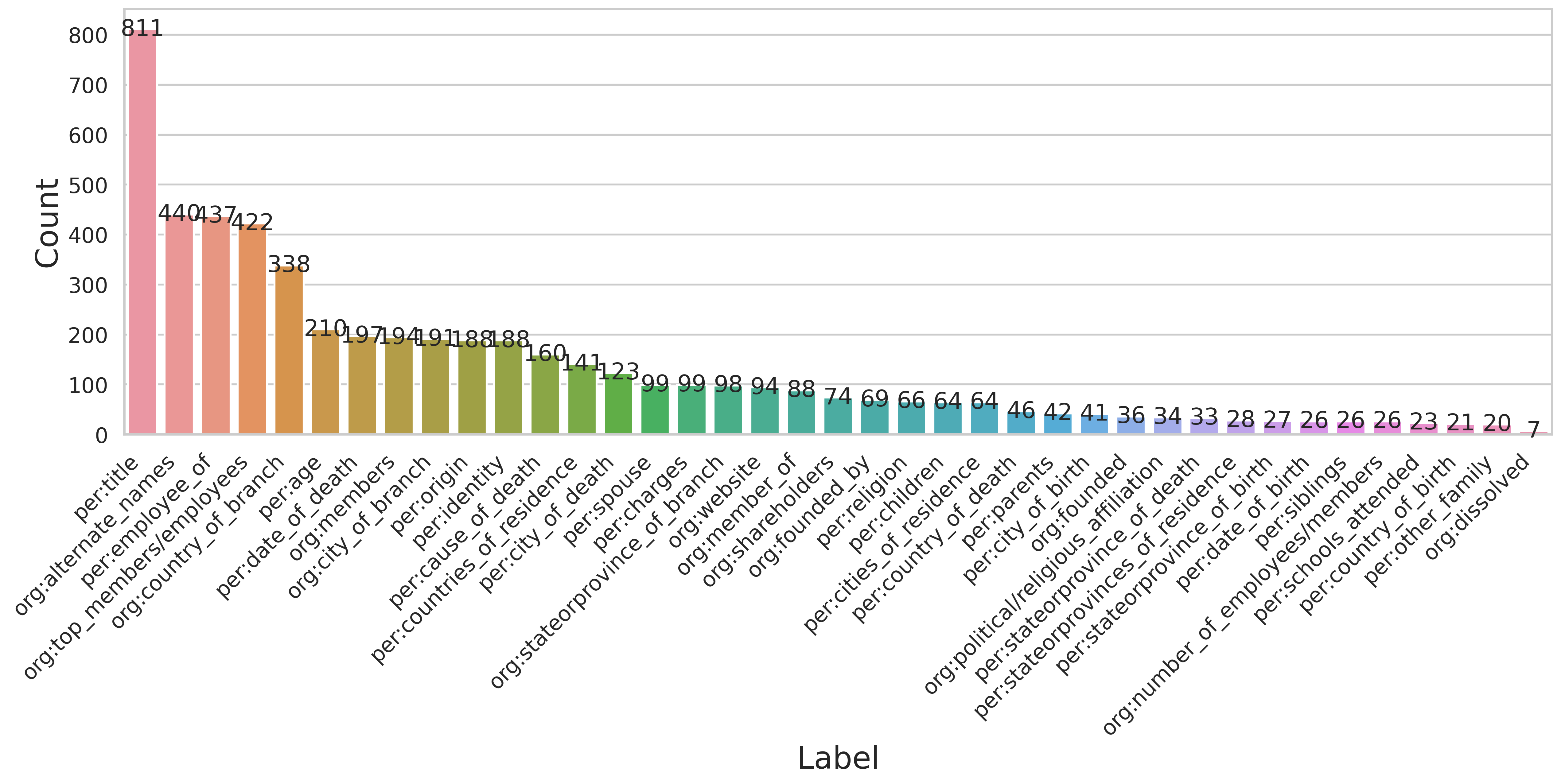}
    \end{minipage}}
    \hfill
    \subfigure[Pronoun Group]{
    \begin{minipage}[]{0,45\linewidth}
    \centering
    \includegraphics[width=3in]{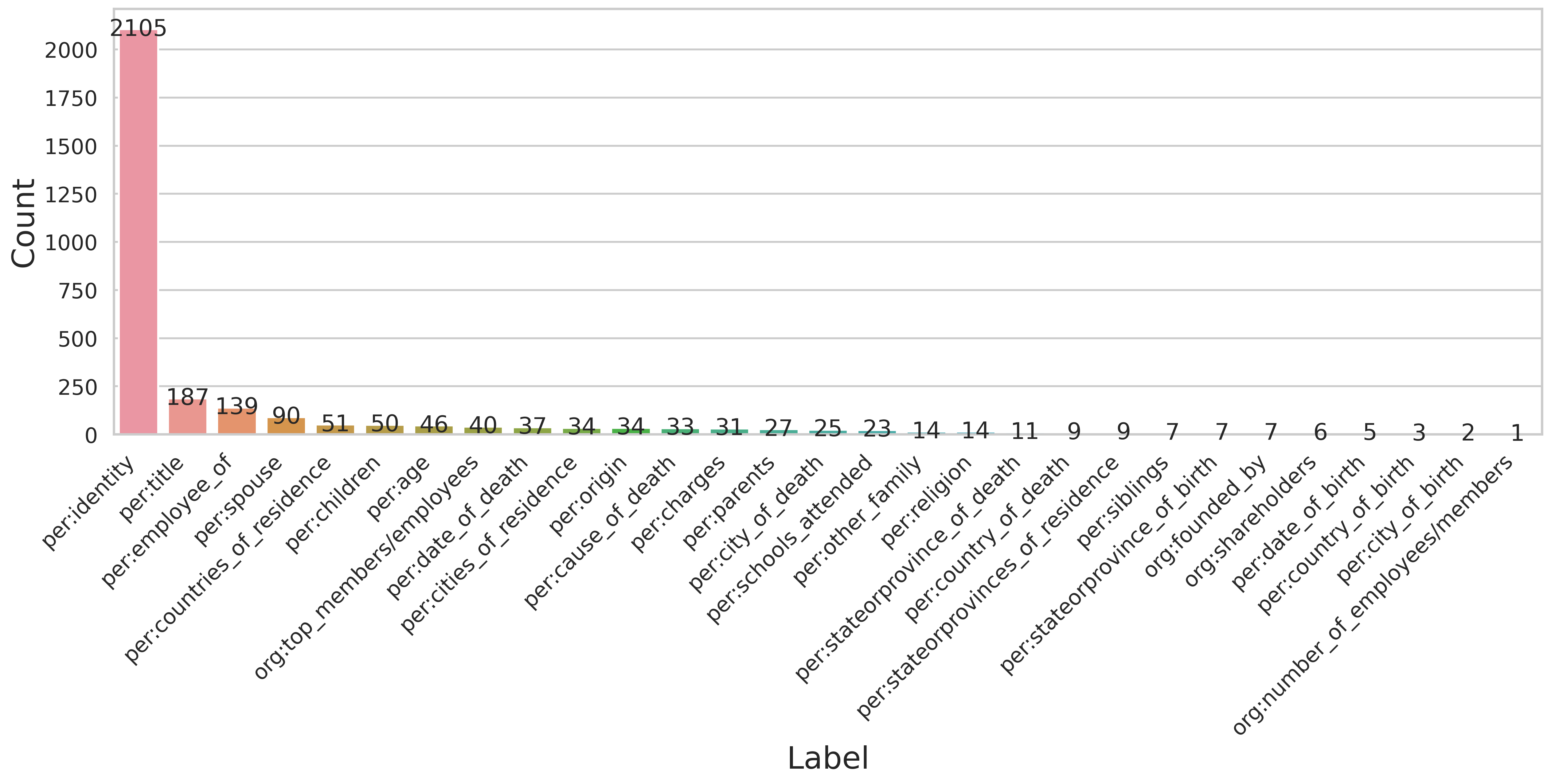}
    \end{minipage}
    } 
    \caption{Lable distribution  of Re-TACRED in two special groups.}
    \label{fig:group}
\end{figure*}

\paragraph{Group Results}
\label{app:group_experiments}
In the Re-TACRED dataset, entities in some instances are represented as pronouns. To assess the efficacy of GMRs in such cases, we divided the data into two distinct groups: Mention Group and Pronoun Group. The division is based on whether the entities within each instance included pronouns. Figure~\ref{fig:group} shows the label distribution of each group. To investigate the effects of GMRs for each group, we re-evaluated each group based on BERT$_{\texttt{base}}$, 
% using the Re-TACRED Settings in Table~\ref{tab:hyperparameter}. 
The results shown in Figure~\ref{tab:group} reveal the positive effects of both types of GMRs within both the Mention Group and the Pronoun Group. 
Specifically, in the Pronoun Group, entities are represented by pronouns, presenting a greater challenge for semantic reasoning due to the lack of explicit semantic references. However, it is worth noting that a significant portion of examples in the Pronoun Group belong to the ``per:identit'' relation, leading to higher scores for the model in this particular group.

\begin{table}[t]
\centering
\scalebox{0.90}{
\begin{tabular}{l|cc}
\toprule
\multirow{2}{*}{\textbf{GMR}} & \multicolumn{2}{c}{\textbf{Group}} \\ \cline{2-3} 
                              & \textbf{Mention} & \textbf{Pronoun} \\ \midrule
\textbf{w/o GMR}                   & 85.86±0.23      & 90.72±0.36       \\ 
\textbf{w/ UD}                   & 86.18±0.25      & 90.93±0.31       \\ 
\textbf{w/ DM}                   & 85.93±0.42      & 91.22±0.20       \\ 
\bottomrule
\end{tabular}}
\caption{Results comparison of each group.}
\label{tab:group}
\end{table}

% \begin{figure*}[t]
%     \centering
%     \subfigure[Re-TACRED]{
%     \begin{minipage}[]{0.31\linewidth}
%     \centering
%     \includegraphics[width=1.7in]{figure/parser/parser_Re_TACRED.png}
%     \end{minipage}
%     } 
%     \hfill
%     \subfigure[SemEval]{
%     \begin{minipage}[]{0.31\linewidth}
%     \centering
%     \includegraphics[width=1.7in]{figure/parser/parser_SemEval.png}
%     \end{minipage}
%     }
%     \hfill
%     \subfigure[FinRE]{
%     \begin{minipage}[]{0.31\linewidth}
%     \centering
%     \includegraphics[width=1.7in]{figure/parser/parser_FinRE.png}
%     \end{minipage}
%     } 
%     % \subfigure[SanWen]{
%     % \begin{minipage}[]{0.22\linewidth}
%     % \centering
%     % \includegraphics[width=1.5in]{figure/parser/parser_SanWen.png}
%     % \end{minipage}
%     % }

%     \caption{Dev F1 score comparison on four datasets with different parsers. The results in each parser include all related layer settings. The dashed lines represent baselines where no graph is encoded}
%     \label{fig:parsers}
% \end{figure*}

\paragraph{Significant Test}  To demonstrate the validity of our framework and the significance of each GMR, we conduct significance tests on the English datasets, which involve multiple base models, including BERT~\cite{devlin-etal-2019-bert}, RoBERTa~\cite{liu2019roberta}, and DeBERTa~\cite{he2020deberta}.
Specifically, we compared all pairs of models based on five random seeds each using ASO~\cite{dror-etal-2019-deep} with a confidence level of 
$\alpha = 0.05$ (before adjusting for all pair-wise comparisons using the Bonferroni correction) by using \textit{deepsig}~\cite{ulmer2022deep}. Almost stochastic 
dominance ($\epsilon_\text{min} < \tau$ with $\tau = 0.2$, recommended by~\citet{ulmer2022deep}) is indicated in Table~\ref{tab:test}. $**$ indicates that our method with UD or DM is significantly better than the baseline. The significant testing results can support our arguments in the paper.

\begin{table}[t]
\centering
\scalebox{0.65}{
\begin{tabular}{@{}llll@{}}
\toprule
\textbf{Model} & \textbf{Encoder} & \textbf{TAC.}         & \textbf{Sem.}         \\ \midrule
IRC            & {BERT}$_{base}$        & 87.66±0.19            & 91.22±0.11            \\
IRC+UD         & {BERT}$_{base}$        & 88.38±0.19 (↑0.72) ** & 91.63±0.24 (↑0.41) ** \\
IRC+DM         & {BERT}$_{base}$        & 88.69±0.21 (↑1.03) ** & 91.58±0.25 (↑0.36) ** \\ \midrule
IRC            & {RoBERTa}$_{base}$     & 88.65±0.23            & 90.62±0.15            \\
IRC+UD         & {RoBERTa}$_{base}$     & 89.08±0.18 (↑0.43) ** & 91.07±0.15 (↑0.45) ** \\
IRC+DM         & {RoBERTa}$_{base}$     & 89.16±0.18 (↑0.17) ** & 91.11±0.13 (↑0.49) ** \\ \midrule
IRC            & {DeBERTa}$_{base}$     & 88.99±0.24            & 91.34±0.11            \\
IRC+UD         & {DeBERTa}$_{base}$     & 89.59±0.22 (↑0.94) ** & 91.70±0.25 (↑0.36) ** \\
IRC+DM         & {DeBERTa}$_{base}$     & 89.25±0.18 (↑0.26) ** & 91.74±0.27 (↑0.40) ** \\ \bottomrule
\end{tabular}}
\caption{Significance Test Results on English Datasets: The table displays average performance and standard deviation computed from five random seeds. Significant results are denoted by $**$. }
\label{tab:test}
\end{table}

% \begin{figure*}
%     \centering
%     \subfigure[Re-TACRED]{
%     \begin{minipage}[]{0.3\linewidth}
%     \centering
%     \includegraphics[width=2in]{figure/random/BERT_ReTACRED.png}
%     \end{minipage}} 
%     \subfigure[SemEval]{
%     \begin{minipage}[]{0.3\linewidth}
%     \centering
%     \includegraphics[width=2in]{figure/random/BERT_SemEval.png}
%     \end{minipage}
%     } 
%     \subfigure[FinRE]{
%     \begin{minipage}[]{0.3\linewidth}
%     \centering
%     \includegraphics[width=2in]{figure/random/BERT_FinRE.png}
%     \end{minipage}
%     }

%     \caption{Dev F1 with different GMR.}
% \end{figure*}

\begin{figure*}[t]
    \centering
    \subfigure[Re-TACRED]{
    \begin{minipage}[]{0.31\linewidth}
    \centering
    \includegraphics[width=1.7in]{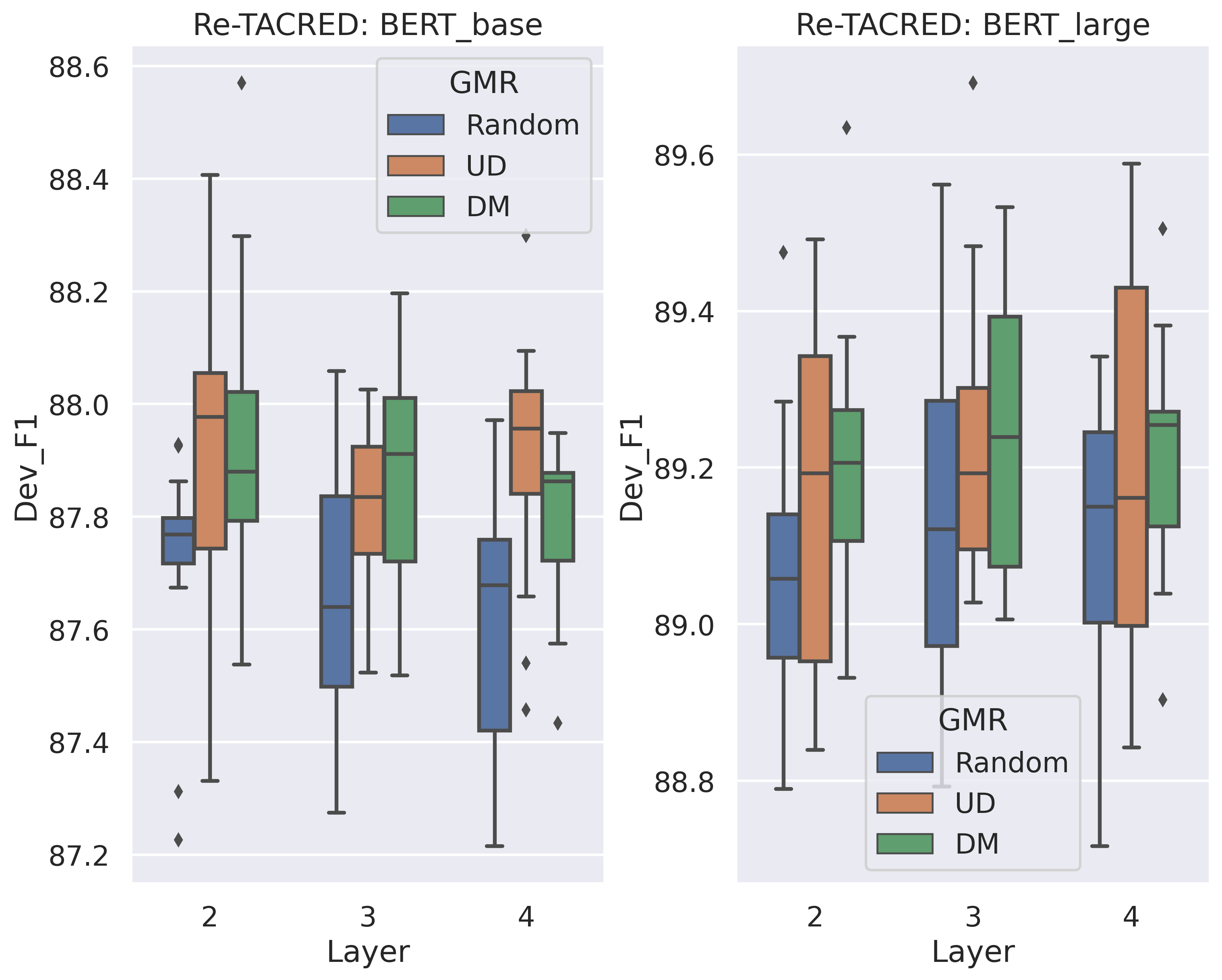}
    \end{minipage}} 
    \hfill
    \subfigure[SemEval]{
    \begin{minipage}[]{0.31\linewidth}
    \centering
    \includegraphics[width=1.7in]{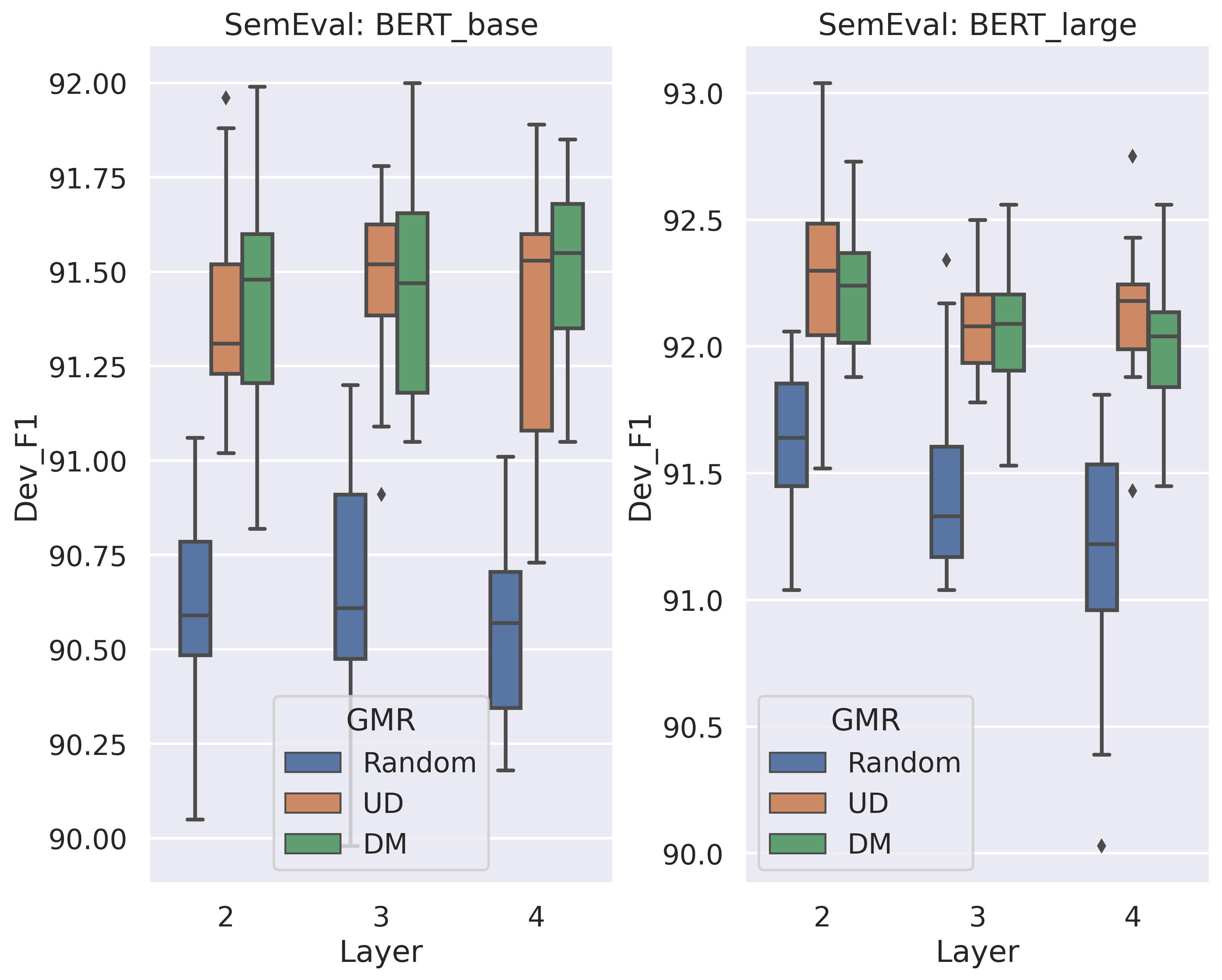}
    \end{minipage}
    } 
    \hfill
    \subfigure[FinRE]{
    \begin{minipage}[]{0.31\linewidth}
    \centering
    \includegraphics[width=1.7in]{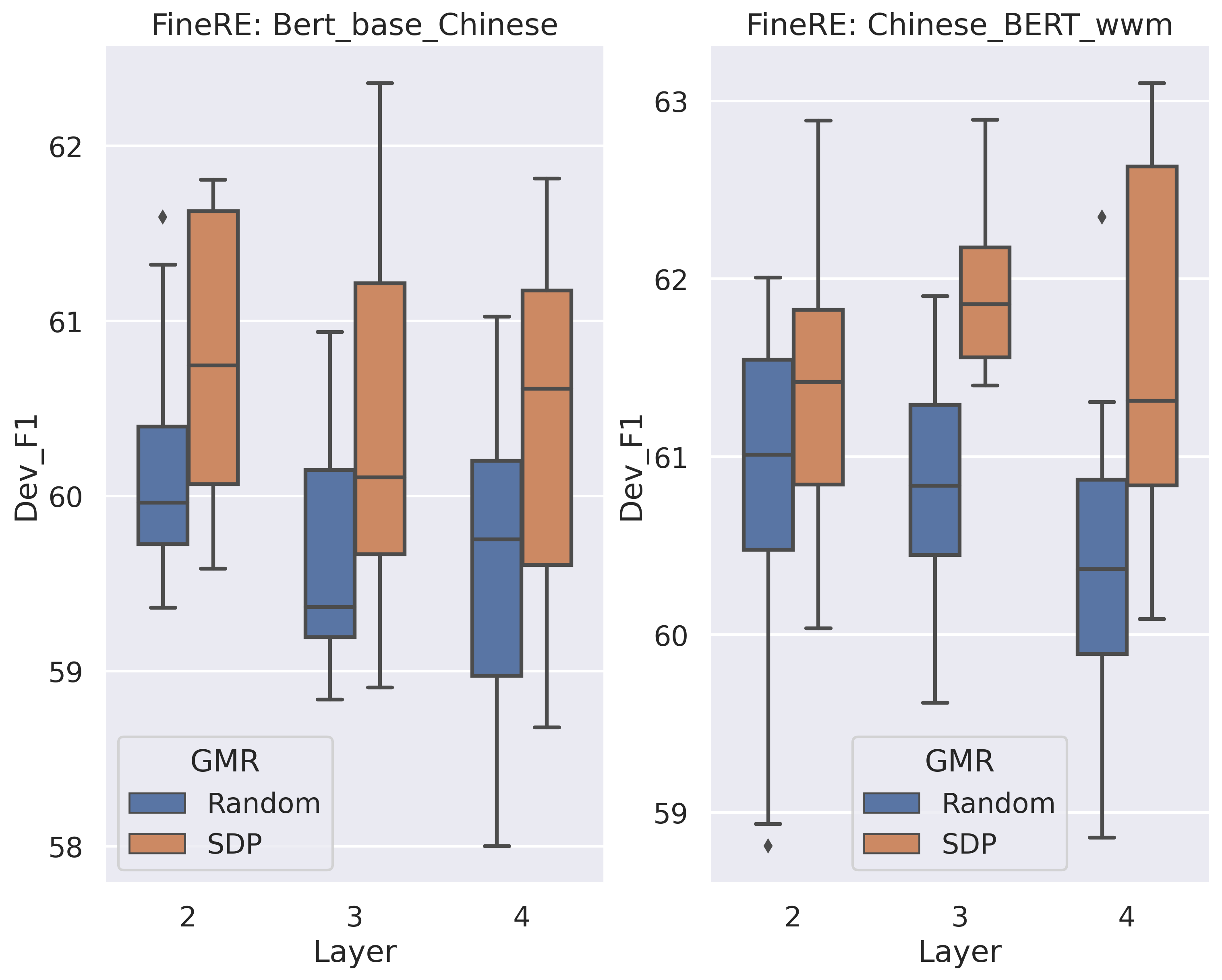}
    \end{minipage}
    }
    % \subfigure[SanWen]{
    % \begin{minipage}[]{0.23\linewidth}
    % \centering
    % \includegraphics[width=1.5in]{figure/random/BERT_SanWen.png}
    % \end{minipage}
    % }

    \caption{Dev F1 score comparison with different GMR frameworks and random graphs.The results in each GMR come from all relevant parsers.}
    \label{fig:GMRs}
\end{figure*}

% \begin{figure*}
%     \centering
%     \subfigure[Re-TACRED]{
%     \begin{minipage}[]{0.3\linewidth}
%     \centering
%     \includegraphics[width=1.5in]{figure/parser/parser_Re_TACRED.png}
%     \end{minipage}
%     } 
%     \subfigure[SemEval]{
%     \begin{minipage}[]{0.3\linewidth}
%     \centering
%     \includegraphics[width=1.5in]{figure/parser/parser_SemEval.png}
%     \end{minipage}
%     }
%     \subfigure[FinRE]{
%     \begin{minipage}[]{0.3\linewidth}
%     \centering
%     \includegraphics[width=1.5in]{figure/parser/parser_FinRE.png}
%     \end{minipage}
%     } 
%     % \subfigure[SanWen]{
%     % \begin{minipage}[]{0.23\linewidth}
%     % \centering
%     % \includegraphics[width=1.5in]{figure/parser/parser_SanWen.png}
%     % \end{minipage}
%     % }
        
%     \caption{Dev F1 on four datasets with different parsers.}
%     \label{fig:GMR}
% \end{figure*}

\begin{figure*}[t]
    \centering
    \subfigure[Re-TACRED]{
    \begin{minipage}[]{0.31\linewidth}
    \centering
    \includegraphics[width=1.7in]{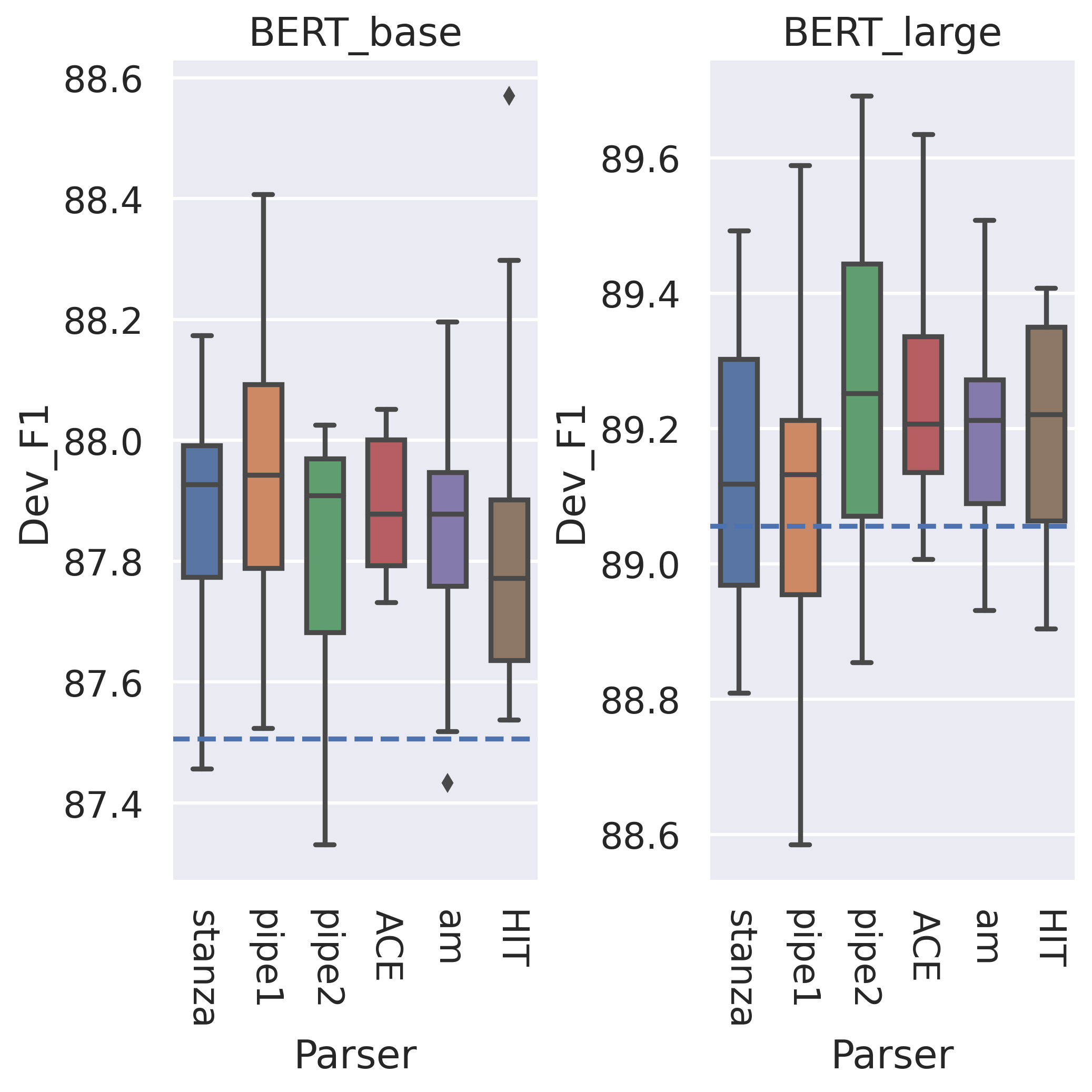}
    \end{minipage}
    } 
    \hfill
    \subfigure[SemEval]{
    \begin{minipage}[]{0.31\linewidth}
    \centering
    \includegraphics[width=1.7in]{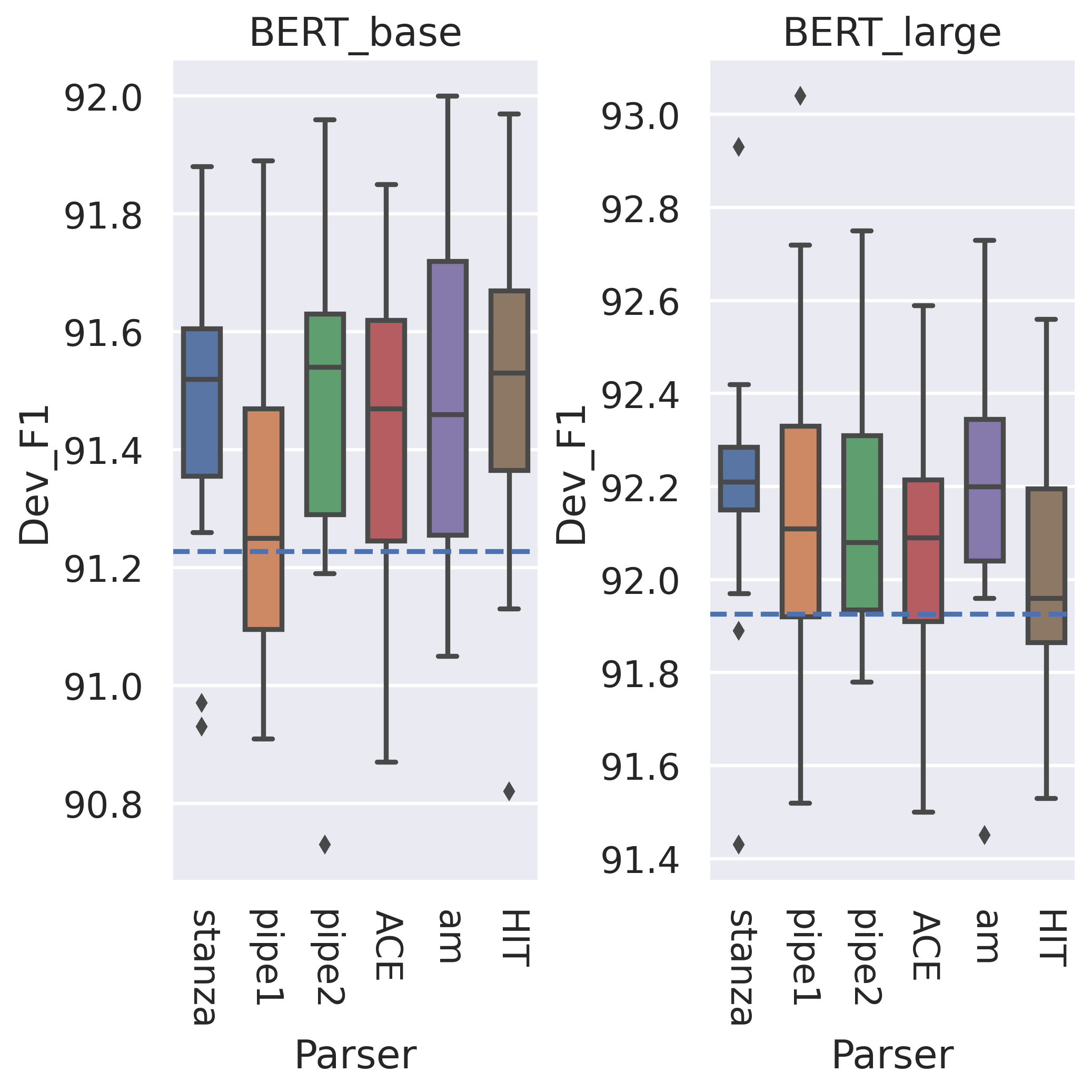}
    \end{minipage}
    }
    \hfill
    \subfigure[FinRE]{
    \begin{minipage}[]{0.31\linewidth}
    \centering
    \includegraphics[width=1.7in]{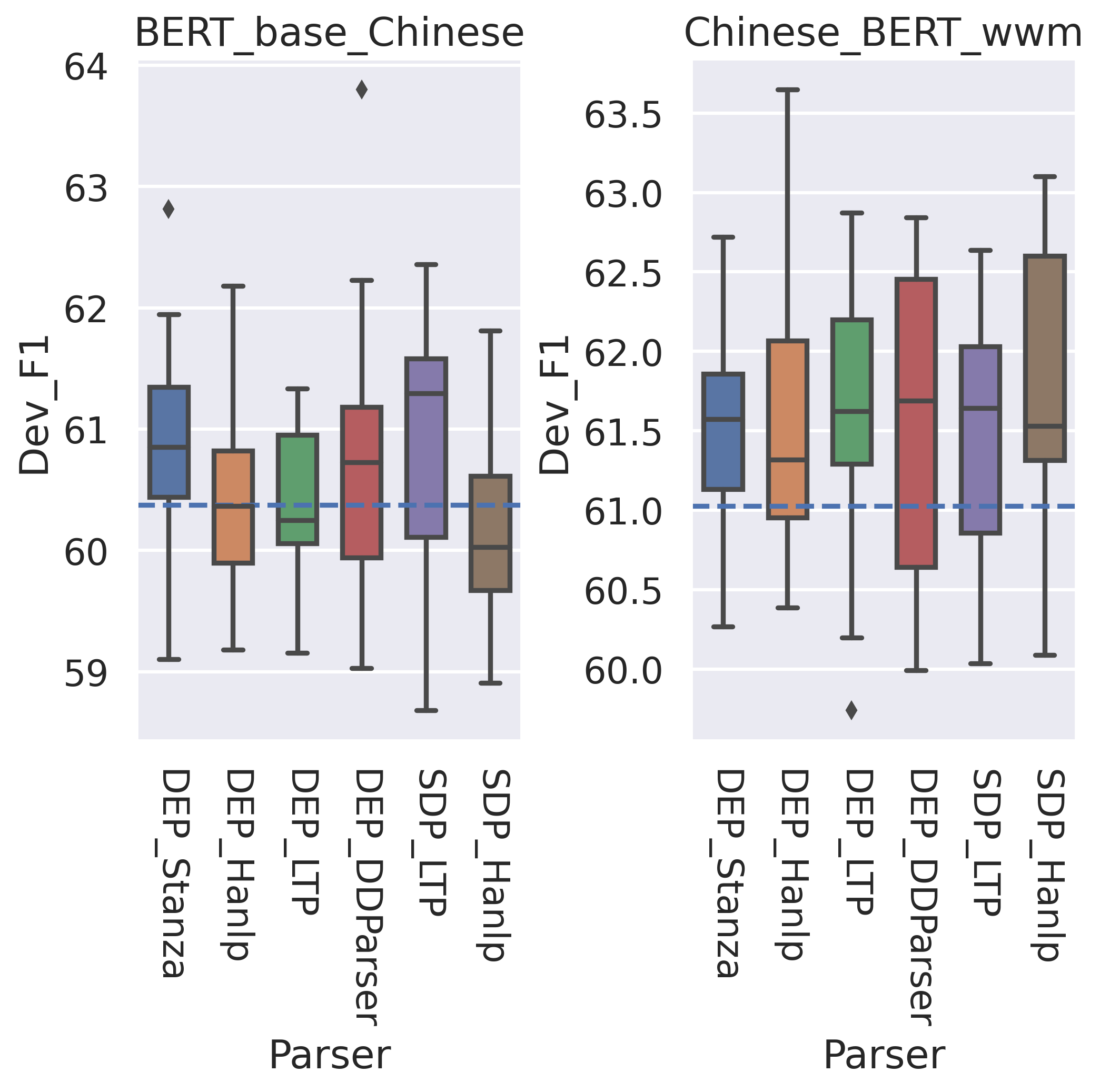}
    \end{minipage}
    } 
    % \subfigure[SanWen]{
    % \begin{minipage}[]{0.22\linewidth}
    % \centering
    % \includegraphics[width=1.5in]{figure/parser/parser_SanWen.png}
    % \end{minipage}
    % }

    \caption{Dev F1 score comparison on four datasets with different parsers. The results in each parser include all related layer settings. The dashed lines represent baselines where no graph is encoded}
    \label{fig:parsers}
\end{figure*}

\section{Analysis}

\subsection{Analysis by GMR Framework}\label{sec:analysisGMR}
% do a statistics

We show the experimental results distribution of each GMR on three datasets in Figure~\ref{fig:GMRs},\footnote{For Chinese datasets, only the comparison with the semantic structure SDP is conducted since there is no standardized annotation for syntactic GMR.} in which the results come from all relevant parsers. 
To verify whether the performance improvement is due to the information provided by the syntactic or semantic structure, we run a control experiment using random graphs. From the results, we find that 1) Regardless of English or Chinese datasets, utilizing GMRs with semantic or syntactic structures consistently yields positive effects when compared to random graphs. 2) With an increase in the number of graph encoder layers, the presence of a random graph can lead to the propagation of more noise information, resulting in a further decline in performance. 3) The decision contribution of GMRs is affected by the sequence encoders.  In the stronger sequence encoders (\texttt{BERT}$_{\texttt{large}}$ and \texttt{Chinese-BERT-wwm}), the contribution of GMRs is weaker. This may be because BERT can encode syntactic information and shows a preference for UD~\cite{kulmizev-etal-2020-neural}.

\subsection{Analysis by Parser Performance}\label{sec:analysisparser}
To assess the impact of each parser on the RC task, we present their performances in Figure~\ref{fig:parsers}, where each parser's samples encompass all relevant layers. This serves as an extrinsic evaluation of the parsers~\cite{fares-etal-2018-2018}, contrasting with the intrinsic evaluation reported in Table~\ref{tab:parser_performance} for parsing benchmarks. In English datasets, UDPipe2 for parsing UD and AM-parser for parsing DM consistently outperform the baseline across various setups involving different sequence encoders and datasets. This observation aligns generally with Table~\ref{tab:parser_performance}, highlighting the crucial role of parser quality in ensuring performance and stability of downstream tasks. However, for Chinese datasets, apart from Stanza, which exhibits a clear advantage on the FinRE dataset, none of the other parsers consistently provide a stable positive effect. This suggests ample room for further development of Chinese parsers.

\subsection{Analysis by Graph Encoders}

\begin{figure}[t]
    \centering
    \includegraphics[width=0.8\linewidth]{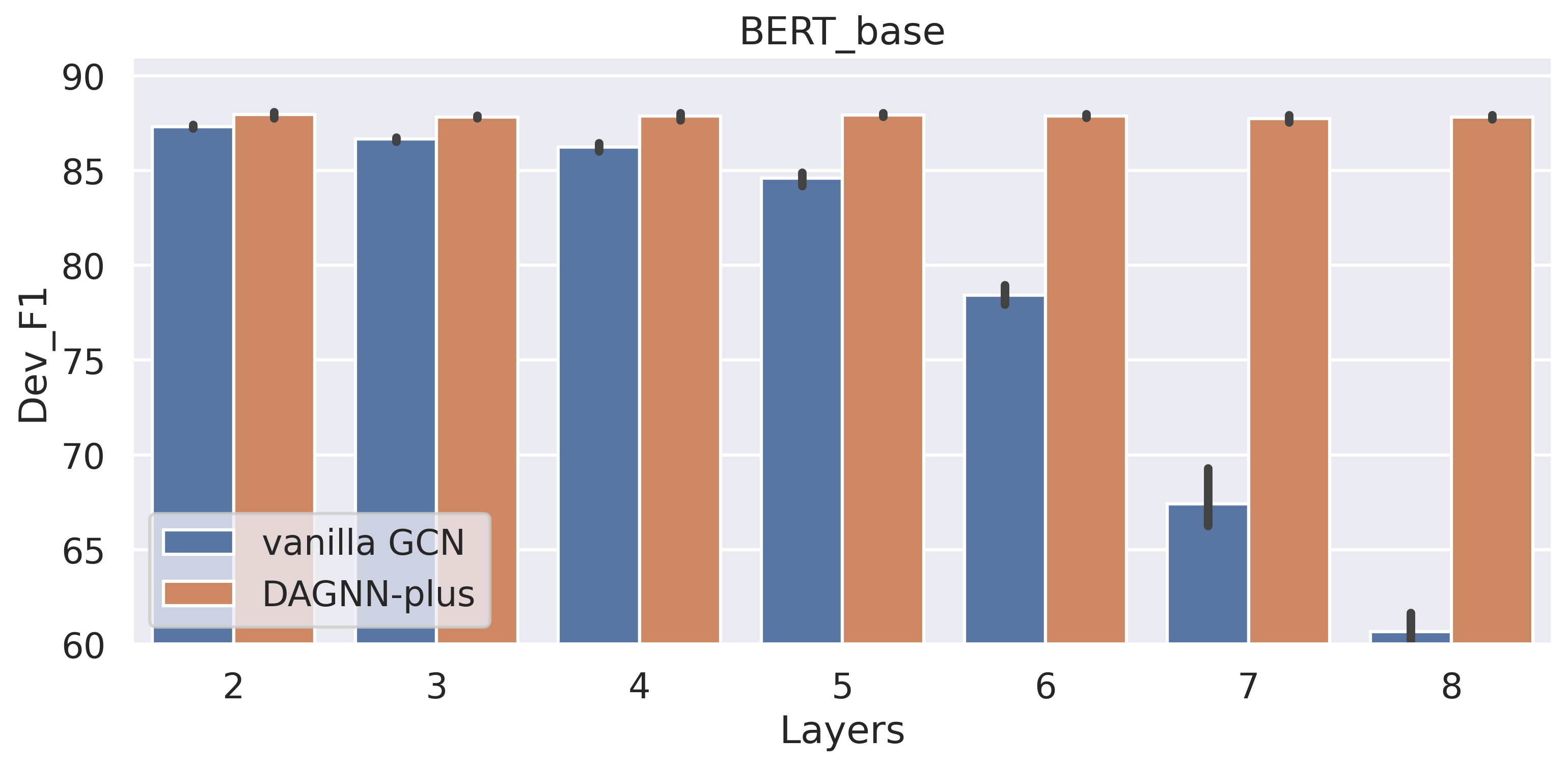}
    \caption{Performance comparison with BERT$_{\texttt{base}}$ and different number of layers among vanilla GCN and DAGNN-plus experimented on English Re-TACRED dataset. We train 5 models with different random seeds for each layer setting and average the results.}
    \label{fig:GE_layer}
\end{figure}

% \begin{table}[]
% \resizebox{\linewidth}{!}{
% \begin{tabular}{lrrrrrrrr}
% \toprule
%             & 2 & 3 & 4 & 5 & 6 & 7 & 8 & TC. \\
% \midrule
% vallian GCN & 2$\text{d}^2$ & 3$\text{d}^2$ & 4$\text{d}^2$ & 5$\text{d}^2$ & 6$\text{d}^2$ & 7$\text{d}^2$ & 8$\text{d}^2$ & O($\text{ld}^2$)   \\
% DAGNN-plus  & d & d & d & d & d & d & d & O($\text{d}$)  \\
% \bottomrule
% \end{tabular}}
% \caption{}
% \label{tab:time}
% \end{table}

To demonstrate the superiority of our proposed DAGNN-plus, we compare the performance with different layers between it and vanilla GCN on the English Re-TACRED dataset, using the sequence encoder \texttt{BERT}$_{\texttt{base}}$. 
As shown in Figure~\ref{fig:GE_layer}, the performance of vanilla GCN decreases with depth due to the loss of the contextual information originally learned by sequence encoders.
However, DAGNN-plus can capture new dependency information while preserving the original context information.
This shows the importance of balancing contextual semantic information and dependency information in neural networks for injecting GMRs.
Furthermore, a vanilla GCN requires $Ld^2$ additional parameters (with $L$ being the graph encoder layer dimension), while DAGNN-plus requires only $d$, being more parameter efficient.

% \subsection{Discussion about Languages and Datasets}

\section{Discussion and Conclusion}
In this paper, we introduce DAGNN-plus, a simple and parameter-efficient method, to systematically investigate the influence of graph meaning representations (GMRs) and their associated parsers on relational classification tasks. Our experimental analysis encompasses both Chinese and English languages, as well as general-domain and literary-domain datasets.
% Unlike \citet{10138896}, which state that semantic representation outperforms syntactic representation in emotion classification, our study
% We do not find a significant difference between syntactic and semantic GMRs. 
Both syntactic and semantic GMRs can effectively improve the performance of RC except on the challenging Chinese literary dataset. A high-quality parser is instrumental in ensuring the performance and robustness of the downstream task of RC. This highlights the significance of ongoing research in the field of parsers.

GMRs exhibit limited effectiveness in Chinese literary text for various reasons. Firstly, most parsers are trained and evaluated using general-domain text, making it difficult for them to accurately parse out-of-domain literary works~\cite{salomoni2017dependency, hershcovich-etal-2017-transition, adelmann2018evaluation, stymne-etal-2023-parser}. Additionally, deep parsing in Chinese is more challenging than in English due to factors like the scarcity of syntactic constraints on Chinese verbs, pro-drop phenomena, and ambiguous constructions~\cite{yu-etal-2011-analysis, kummerfeld-etal-2013-empirical}. Despite the presence of advanced Chinese parsers, the lack of standardized annotation and limited evaluation resources hinder their development. Moreover, Chinese literary language presents additional challenges, including complex linguistic structures, limited annotated training data, domain-specific knowledge requirements, and the absence of explicit graph structures. These factors, combined with the need to handle poetic and metaphorical expressions, call for specialized approaches and benchmarks to effectively parse GMRs in Chinese literary language.

\section*{Limitations}
While we experiment with parsers of different levels of performance, we have not experimented with gold syntactic or semantic parsers, as they are not available for the datasets we experimented with (or for other RC datasets). As shown by \citet[\textit{inter alia}]{sachan-etal-2021-syntax} for another task (SRL), gold parses can improve results even beyond SOTA parsers. This ideal experiment, if it could be done for RC, would shed light on the best possible performance achievable with the GNN approaches we investigate.

% \section*{Ethics Statement}

% \section*{Acknowledgements}

\bibliography{anthology,custom}
\bibliographystyle{acl_natbib}

\end{document}